\documentclass{article}

\usepackage{arxiv}

\usepackage{amsmath, amsthm, amssymb, graphicx, hyperref, etoolbox, longtable, colonequals}

\makeatletter
\let\@nodottedtocline\@dottedtocline
\patchcmd{\@nodottedtocline}{\hbox{.}}{\hbox{}}{}{}
\patchcmd{\@nodottedtocline}{\normalcolor #5}{\normalcolor}{}{}
\newcommand*\l@sectionsubtitle{\@nodottedtocline{1}{0em}{1.5em}}
\makeatother

\usepackage{imakeidx}
\makeindex[title=Index,columns=3]
\indexsetup{noclearpage}

\usepackage{tikz}
\usetikzlibrary{calc,arrows.meta}

\theoremstyle{definition}

\newcommand\C{\mathbf{C}}

\newcommand\cofib\rightarrowtail

\newcommand\Hom{\textup{Hom}}

\newcommand\mdel[1]{}

\newcommand\Set{\textup{Set}}

\newcommand\sObj{\textup{sObj}}

\newcommand\sSet{\textup{sSet}}

\renewcommand\geq\geqslant
\renewcommand\leq\leqslant

\usepackage[utf8]{inputenc} 
\usepackage[T1]{fontenc}    
\usepackage{hyperref}       
\usepackage{url}            
\usepackage{booktabs}       
\usepackage{amsfonts}       
\usepackage{nicefrac}       
\usepackage{microtype}      
\usepackage{lipsum}		
\usepackage{graphicx}
\usepackage{natbib}
\usepackage{doi}
\usepackage{tikz-cd}
\usepackage{mathtools}

\usepackage{amsmath}
\usepackage{amssymb}
\usepackage{amsthm}
\usepackage[boxruled]{algorithm2e}

\newtheorem{theorem}{Theorem}
\newtheorem{definition}{Definition}

\newtheorem{example}{Example}

\newcommand{\CI}{\mathrel{\perp\mspace{-10mu}\perp}}

\title{Unifying Causal Inference and Reinforcement Learning using Higher-Order Category Theory \thanks{Draft under revision. Comments welcome.} }

\author{ Sridhar Mahadevan \\
	Adobe Research and University of Massachusetts, Amherst\\
	\texttt{smahadev@adobe.com, mahadeva@umass.edu}
}

\hypersetup{
pdftitle={A template for the arxiv style},
pdfsubject={q-bio.NC, q-bio.QM},
pdfauthor={David S.~Hippocampus, Elias D.~Striatum},
pdfkeywords={First keyword, Second keyword, More},
}

\begin{document}
\maketitle

\begin{abstract}
We present a  unified  formalism for structure discovery of causal models and predictive state representation (PSR) models in reinforcement learning (RL) using higher-order category theory. Specifically, we model structure discovery in both settings using simplicial objects, contravariant functors from the category of ordinal numbers into any category. Fragments of causal models that are equivalent under conditional independence -- defined as causal horns -- as well as subsequences of potential tests in a predictive state representation -- defined as predictive horns -- are both special cases of horns of a simplicial object,  subsets resulting from the removal of the interior and the face opposite a particular vertex. Latent structure discovery in both settings involve the same fundamental mathematical problem of finding extensions of horns of simplicial objects through solving lifting problems in commutative diagrams, and exploiting weak homotopies that define higher-order symmetries. Solutions to the problem of filling ``inner" vs ``outer" horns leads to various notions of higher-order categories, from Kan complexes to quasicategories and $\infty$-category theory. We define the abstract problem of structure discovery in both settings in terms of adjoint functors between the category of universal causal models or universal decision models and its simplicial object representation.   In general, the left adjoint functor from a simplicial object $X$ to a category ${\cal C}$ is lossy, preserving only relationships up to a certain order defined by homotopical equivalences. In contrast, the right adjoint defining the nerve of a category constructs a lossless encoding of a category as a simplicial object. 
\end{abstract}

\keywords{AI \and Category Theory \and Causal Inference  \and Simplicial Objects \and Machine Learning \and Statistics}

\section{Introduction}

\begin{figure}[p] 
\centering
\vskip 0.1in
\begin{minipage}{0.9\textwidth}
\includegraphics[scale=0.38]{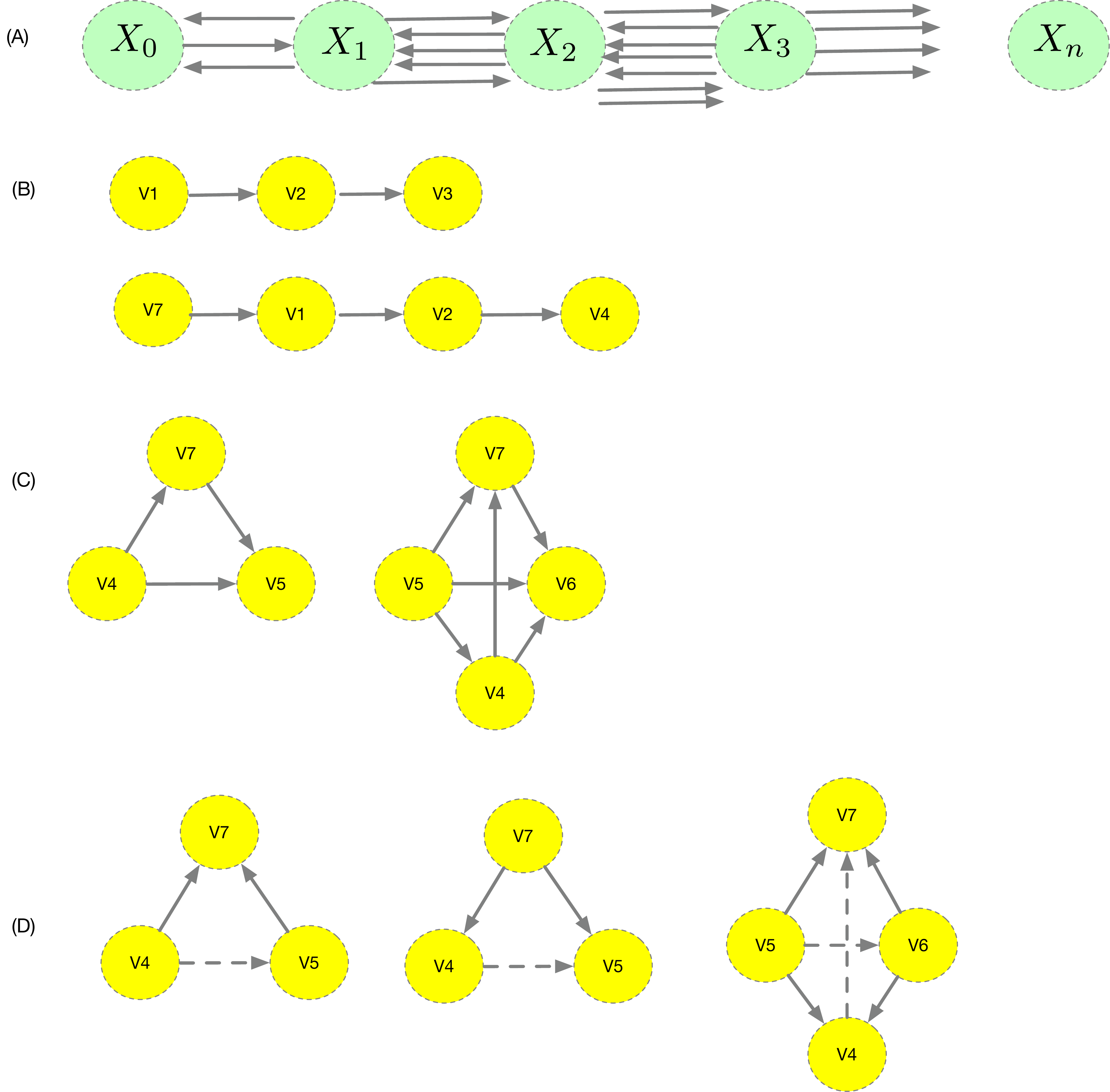}
\end{minipage}
\caption{(A): In this paper, we unify structure prediction in causal inference and RL in terms of a simplicial object $X_n, n \geq 0$, contravariant functors from ordinal numbers into a category that originated in algebraic topology \citep{may1992simplicial}, but has since become the foundation for higher-order category theory \citep{weakkan,quasicats,kerodon}. The internal structure of $X_n$ involves composable morphisms of dimension $n$. $0$th order objects in $X_0$ involve only identity morphisms from an object to itself. $1$th order objects in $X_1$ define binary arrows between pairs of objects.  Higher-order objects in $X_n, n \geq 2$, induce higher-order relationships including symmetries. Structure discovery in  causal inference and RL is modeled as the extension problems of ``filling horns", where the horn $\Lambda^n_k$ is a subset of a contravariant functor of an $n$-simplex that results from removing its interior and the face  opposite the $k$-vertex. (B): Predictive state horns are partial fragments of potential tests, which can faithfully embed a category of dynamical systems into a simplicial object, and can be seen as a special case of the {\em nerve} of a category that defines a fully faithful embedding of arbitrary categories in terms of simplicial objects.   (C): Causal ``inner" horns, fragments of causal models, define a {\em quasicategory} \citep{quasicats}, a  simplicial object where composition of oriented $n$-simplices involves solving a lifting problem in a commutative diagram under a weak homotopy. (D) ``Outer" horns define more general compositional structures that are analyzed in  $\infty$-category theory \citep{kerodon}.  }
\label{highercat}
 \end{figure} 

Causal inference \citep{pearl:causalitybook,rubin-book,spirtes:book} and predictive state representations (PSRs) \citep{singh-uai04} in reinforcement learning \citep{DBLP:books/lib/SuttonB98}, whose roots go back to earlier work on subspace identification in linear systems \citep{overschee} and even earlier work on algebraic theories of context-free languages \cite{CHOMSKY1963118} and algebraic automata theory \citep{GIVEON1968346},  both involve structure discovery of a latent variable model through interventions. The use of superficially dissimilar representations --  directed acyclic graphs (DAGs)  \citep{pearl:bnets-book},  hybrid undirected and directed graphs \citep{lauritzen:chain} and hyperedge graphs \citep{hedge,mdag} in causal inference, versus Hankel matrix and Hilbert space embeddings  of dynamical systems -- have long obscured their deeper connections. Structure discovery in causal inference and PSRs both involve the determination of a latent structure, which is directional at lower orders, but homotopy equivalences  at higher orders induce symmetries. In particular, causal inference involves determining a structure, such as a DAG that encodes direct causal effects between a pair of objects, but multiple DAG models are equivalent because of symmetries induced by conditional independences \citep{DBLP:journals/amai/Dawid01,STUDENY2010573} and  correlations induced by latent unobservable confounders that are only revealed over higher-order simplices (e.g., DAGs over $n \geq 3$ vertices). PSRs represent ``hidden state" in dynamical systems by constructing a series of tests, a free algebra over paired actions and observation that under certain conditions is guaranteed to faithfully embed a dynamical system without requiring the need to explicitly construct ``belief states", probability distributions over latent states, as in a partially observable Markov decision process (POMDP). The core tests define a minimal left {\bf k}-module over an Abelian group defined as a free algebra over all possible tests,  and induce higher-order symmetries.  

We formulate structure discovery of causal and PSR models in terms of  {\em simplicial objects} (see Figure~\ref{highercat} and Figure~\ref{simpobj}) which were originally introduced as a combinatorial representation in algebraic topology \citep{may1992simplicial}, but have  become the foundation for higher-order category theory \citep{weakkan,quasicats,kerodon}. Formally, simplicial objects \cite{may1992simplicial}  are  contravariant functors $X: \Delta^{o} \rightarrow {\cal C}$ from the category of ordinal numbers, whose objects are $[n] = (0,1, \ldots, n), n \geq 0$, and whose arrows are non-decreasing maps, into an underlying category, such as  a universal causal model \citep{sm:uc}, or a universal decision model \citep{sm:udm}. The $0$th order simplices are objects  in a universal causal model, or  a universal decision model.  The $1$-simplices define arrows encoding  directional relationships between causal or decision objects.   Simplicial objects of higher dimensions $n \geq 2$ encode  higher-order  relationships between groups of objects. 

The general problem of horn filling has been solved in different ways in higher-order category theory, including weak Kan complexes \cite{weakkan}, quasi-categories \citep{quasicats}, and $\infty$-categories \citep{kerodon}. We will review this literature, and then show the connections to structure discovery in causal inference and RL. Structure prediction in causal inference and RL both can be formulated as filling horns of simplicial objects, subsets of functors that result from interventions that removes the interior and the face opposite a given vertex. These interventions can be defined in terms of a sequence of elementary order-preserving morphisms in the category of ordinal numbers $\Delta$, whose objects are the ordinals $[n] = \{0, 1, \ldots, n \}$ and whose arrows are defined as compositions of elementary injections $d_i: [n-1] \rightarrow [n]$ and elementary surjections $s_i: [n] \rightarrow [n-1]$ (see Figure~\ref{simpobj}). In particular, the units in a causal study or the states of an underlying dynamical system are $x, y, \ldots \in X_0$, the $0$-simplices of $X$.  If $x, y \in X_0$, and $f \in X_1$, we denote a potential causal effect or action in a dynamical system by the arrow $f: x \rightarrow y$ to denote that $x = d_1 f$ and $y = d_0 f$. Here,  $d_i$ is an elementary injection $d_i: [n-1] \rightarrow [n]$ that skips element $i$.  The set of potential causal influences or actions between $x$ and $y$ is denoted as $X_1(x,y)$. The degenerate $s_0 x$ denotes the unit morphism (a cycle) mapping ${\bf 1}_x: x \rightarrow x$, where $s_i: [n] \rightarrow [n=1]$ is the elementary surjection that repeats element $i \in [n]$.  Higher-order simplices $X_i, i \geq 2$ are used to capture higher-order causal effects or action transitions among groups of causal objects or dynamical system states of size $n > 2$. Note in Figure~\ref{highercat}, there is one arrow out of $X_0$, denoting the self-mapping from an object to itself. The two arrows from $X_1$ to $X_0$ represent the head and tail of each morphism in $X_1$. There are in general $n+1$ arrows in and out of $X_n$, for $n > 0$. For example, $X_2$ has three arrows going back into $X_1$, because those are the three arrows that are contained in the $2$-simplex. 

\section{Simplicial Objects} 

 Simplicial objects have long been a foundation for algebraic topology \citep{may1999concise,may1992simplicial}, and  more recently higher-order category theory \citep{weakkan,quasicats,kerodon}. The category $\Delta$ has non-empty ordinals $[n] = \{0, 1, \ldots, n]$ as objects, and order-preserving maps $[m] \rightarrow [n]$ as arrows. An important property in $\Delta$ is that any mapping is decomposable as a composition of an injective and a surjective mapping,  each of which is decomposable into a sequence of elementary injections $d_i: [n] \rightarrow [n+1]$, which omits $i \in [n]$, and a sequence of elementary surjections $s_i: [n] \rightarrow [n-1]$, which repeats $i \in [n]$. The fundamental simplex $\Delta([n])$ is the presheaf of all morphisms into $[n]$, that is, the representable functor $\Delta(-, [n])$. \footnote{A representable functor is one that can be faithfully embedded in the category of {\bf Sets}.} 
 
 We will denote simplicial objects over the category of sets by $X[n] = X_n$ to mean the functor $[n] \rightarrow X$, mapping both the objects $[n] = \{0, 1, \ldots, n \}$ as well as its morphisms to a set $X$. Thus, $X_0$ is a set of objects, which are mappings from $[0] = \{ 0 \}$ to $X$, and as $[0]$ contains only one element, each such mapping must pick out a single element of $X$. Note that as the only morphism in $[0]$ is $0 \rightarrow 0$, its image gives the identity mapping on each element in the set $X$. Similarly, $X_1$ is the functor mapping $[1] = \{0, 1 \}$ and its non-identity morphism $0 \rightarrow 1$ to $X$, which defines arrows between elements of $X_0$.  Proceeding, $X[2]$ is the functor mapping $[2] = \{0, 1, 2 \}$, with its non-identity morphisms $0 \rightarrow 1, 1 \rightarrow 2, 0 \rightarrow 2$ into a category, so it picks out ``triangles", or oriented simplices of order $2$. 
 
\begin{figure}[t] 
\centering
\begin{minipage}{0.9\textwidth}
\begin{tabular}{r l l}
Category $\Delta$: & Objects are  the ordinal numbers $[n]=(0,1,\dots,n)$ \\ & \multicolumn{2}{l}{Arrows are non-decreasing  maps $[n]\to[m]$}\\[5pt]
\emph{Elementary coface maps}: & $s_i:[n]\to [n-1]$, repeat $i$ twice in the co-domain& \\
\emph{Elementary codegeneracy maps}: &  $d_i:[n]\to [n+1]$, skips $i$ in the co-domain & \\[5pt]
& Any arrow $[m] \rightarrow [n]$ is composed of elementary face and degeneracy maps: \\
& \multicolumn{2}{c}{
\begin{tikzpicture}[xscale=1,yscale=.7]
\foreach \x\y\namme in {0/1/1c, 0/2/1b, 0/3/1a, 1/0/2d, 1/1/2c, 1/2/2b, 1/3/2a}{
  \coordinate (\namme) at (\x,\y);
  \node[scale=.8] at (\x,\y) {$\bullet$};
}
\foreach \a\b in {1a/2a, 1b/2d, 1c/2d}{
  \draw[->,shorten >=5pt, shorten <=5pt] (\a)--(\b);
}
\begin{scope}[shift={(2.5,0)}]
\foreach \x\y\namme in {0/1/1c, 0/2/1b, 0/3/1a, 1/2/2b, 1/3/2a, 2/1/3c, 2/2/3b, 2/3/3a, 3/0/4d, 3/1/4c, 3/2/4b, 3/3/4a}{
  \coordinate (\namme) at (\x,\y);
  \node[scale=.8] at (\x,\y) {$\bullet$};
}
\foreach \a\b in {1a/2a, 1b/2b, 1c/2b, 2a/3a, 2b/3c, 3a/4a, 3b/4b, 3c/4d}{
  \draw[->,shorten >=5pt, shorten <=5pt] (\a)--(\b);
}
\node at (.5,3.5) {$s_1$};
\node at (1.5,3.5) {$d_1$};
\node at (2.5,3.5) {$d_2$};
\end{scope}
\begin{scope}[shift={(7,0)}]
\foreach \x\y\namme in {0/1/1c, 0/2/1b, 0/3/1a, 1/0/2d, 1/1/2c, 1/2/2b, 1/3/2a, 2/1/3c, 2/2/3b, 2/3/3a, 3/0/4d, 3/1/4c, 3/2/4b, 3/3/4a}{
  \coordinate (\namme) at (\x,\y);
  \node[scale=.8] at (\x,\y) {$\bullet$};
}
\foreach \a\b in {1a/2a, 1b/2c, 1c/2d, 2a/3a, 2b/3b, 2c/3c, 2d/3c, 3a/4a, 3b/4b, 3c/4d}{
  \draw[->,shorten >=5pt, shorten <=5pt] (\a)--(\b);
}
\node at (.5,3.5) {$d_1$};
\node at (1.5,3.5) {$s_2$};
\node at (2.5,3.5) {$d_2$};
\end{scope}
\node at (1.75,2) {is};
\node at (6.25,2) {or};
\end{tikzpicture}
}\\[10pt]
$[n]$: & Objects are numbers $0,1,\dots,n$ \\ & $|\Hom_{[n]}(a,b)| = 1$ iff $a\leqslant b$, else $\emptyset$\\[10pt]
Simplicial set: $\sSet$: & ${\bf Hom}_{\Delta}(\Delta^{op},\Set)$ \\[5pt]
& \multicolumn{2}{l}{Objects are functors $S = \{S_n=S([n])\}_{n\geqslant 0}$ with} \\
& \emph{face maps} $S(s_i):S_n\to S_{n+1}$ & \\
& \emph{degeneracy maps} $S(d_i):S_n\to S_{n-1}$ & \\[5pt]
& \multicolumn{2}{l}{Morphisms $f:S\to T$ are natural transformations} \\[10pt]
Simplicial object: $\sObj$: & ${\bf Hom}_{\Delta}(\Delta^{op},C)$ for $C$ any category
\end{tabular}
\vskip -0.5in
\end{minipage}
\caption{Simplicial objects are contravariant functors from ordinal numbers into any category.} \label{simpobj}
 \end{figure}
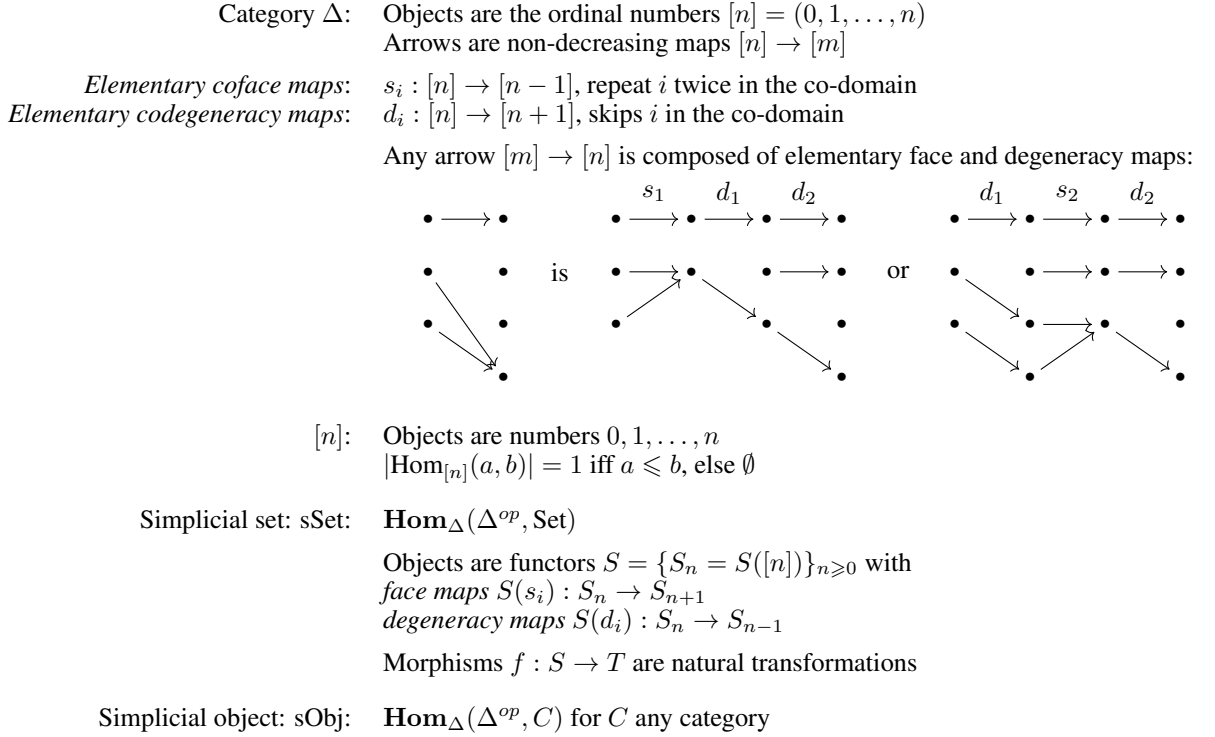

 The Yoneda Lemma \cite{maclane:71} assures us that an $n$-simplex $x \in X_n$ can be identified with the corresponding map $\Delta[n] \rightarrow X$. Every morphism $f: [n] \rightarrow [m]$ in $\Delta$ is functorially mapped to the map $\Delta[m] \rightarrow \Delta[n]$ in ${\cal S}$. Figure~\ref{simpobj}) illustrates simplicial objects, in particular showing that any order-preserving morphism can be decomposed into a sequence of elementary degeneracy and face operators. \footnote{The Latex source for this diagram in is from  {\tt http://homepages.math.uic.edu/$\sim$jlv/seminars/infinitycats/inftycats.tex}.}

Any morphism in the category $\Delta$ can be defined as a sequence of {\em degeneracy} and {\em face} operators, where the degeneracy operator $\delta_i: [n-1] \rightarrow [n], 0 \leq i \leq n$ is defined as: 

\[ 
\delta_i (j)  =
\left\{
	\begin{array}{ll}
		j,  & \mbox{for } \ 0 \leq j \leq i-1 \\
		j+1 & \mbox{for } \  i \leq j \leq n-1 
	\end{array}
\right. \] 

Analogously, the face operator $\sigma_j: [n+1] \rightarrow [n]$ is defined as 

\[ 
\sigma_j (k)  =
\left\{
	\begin{array}{ll}
		j,  & \mbox{for } \ 0 \leq k \leq j \\
		k-1 & \mbox{for } \  j < k \leq n+1 
	\end{array}
\right. \] 

The compositions of these arrows define certain well-known properties \citep{may1992simplicial,richter2020categories}: 

\begin{eqnarray*}
    \delta_j \circ \delta_i &=& \delta_i \circ \delta_{j-1}, \ \ i < j \\
    \sigma_j \circ \sigma_i &=& \sigma_i \circ \sigma_{j+1}, \ \ i \leq j \\ 
    \sigma_j \circ \delta_i (j)  &=&
\left\{
	\begin{array}{ll}
		\sigma_i \circ \sigma_{j+1},  & \mbox{for } \ i < j \\
		1_{[n]} & \mbox{for } \  i = j, j+1 \\ 
		\sigma_{i-1} \circ \sigma_j, \mbox{for} \ i > j + 1
	\end{array}
\right.
\end{eqnarray*}

\begin{definition}
The set of all functors $[n] \rightarrow {\cal C}$ for any category ${\cal C}$ is denoted as ${\cal N}_n({\cal C})$. The image of each abstract face operator $\sigma_j: [n+1] \rightarrow [n]$ under the covariant functor $N_n({\cal C})$ is denoted as $N_n(\sigma_j) = d_j$, and the image of each abstract degeneracy operator $\delta_i: [n-1] \rightarrow [n]$ under the covariant functor $N_n({\cal C})$ is given as $N_n(\delta_i) = s_i$. 
\end{definition}

\subsection{Full and Faithful Embedding of Categories} 

\begin{definition} 
A {\bf covariant functor} $F: {\cal C} \rightarrow {\cal D}$ from category ${\cal C}$ to category ${\cal D}$, and defined as the following: 
\begin{itemize} 
    \item An object $F X$ (sometimes written as $F(x)$) of the category ${\cal D}$ for each object $X$ in category ${\cal C}$.
    \item An  arrow  $F(f): F X \rightarrow F Y$ in category ${\cal D}$ for every arrow  $f: X \rightarrow Y$ in category ${\cal C}$. 
   \item The preservation of identity and composition: $F \ id_X = id_{F X}$ and $(F f) (F g) = F(g \circ f)$ for any composable arrows $f: X \rightarrow Y, g: Y \rightarrow Z$. 
\end{itemize}
\end{definition} 

\begin{definition} 
A {\bf contravariant functor} $F: {\cal C} \rightarrow {\cal D}$ from category ${\cal C}$ to category ${\cal D}$ is defined exactly like the covariant functor, except all the arrows are reversed. In the contravariant functor$F: C^{\mbox{op}} \rightarrow D$, every morphism $f: X \rightarrow Y$ is assigned the reverse morphism $F f: F Y \rightarrow F X$ in category ${\cal D}$. 
\end{definition} 

\begin{itemize} 
\item For every object $X$ in a category ${\cal C}$, there exists a covariant functor ${\cal C}(X, -): {\cal C} \rightarrow {\bf Set}$ that assigns to each object $Z$ in ${\cal C}$ the set of morphisms ${\cal C}(X,Z)$, and to each morphism $f: Y \rightarrow Z$, the pushforward mapping $f_*:{\cal C}(X,Y) \rightarrow {\cal C}(X, Z)$. 

\item For every object $X$ in a category ${\cal C}$, there exists a contravariant functor ${\cal C}(-, X): {\cal C}^{\mbox{op}} \rightarrow {\bf Set}$ that assigns to each object $Z$ in ${\cal C}$ the set of morphisms {\bf Hom}$_{\cal C}(X,Z)$, and to each morphism $f: Y \rightarrow Z$, the pullback mapping $f^*:$ {\bf Hom}$_{\cal C}(Z, X) \rightarrow {\cal C}(Y, X)$. Note how ``contravariance" implies the morphisms in the original category are reversed through the functorial mapping, whereas in covariance, the morphisms are not flipped.
\end{itemize} 

\begin{definition} 
\label{fully-faithful} 
Let ${\cal F}: {\cal C} \rightarrow {\cal D}$ be a functor from category ${\cal C}$ to category ${\cal D}$. If for all arrows $f$ the mapping $f \rightarrow F f$
\begin{itemize}
    \item injective, then the functor ${\cal F}$ is defined to be {\bf faithful}. 
    \item surjective, then the functor ${\cal F}$ is defined to be {\bf full}.  
    \item bijective, then the functor ${\cal F}$ is defined to be {\bf fully faithful}. 
\end{itemize}
\end{definition} 

\begin{definition}
A pair of {\bf adjoint functors} is defined as $F: {\cal C}\rightarrow {\cal D}$ and $G: {\cal D} \rightarrow {\cal C}$, where $F$ is considered the right adjoint, and $G$ is considered the left adjoint. 

\[
        \begin{tikzcd}
            {\cal D} \arrow[r, shift left=1ex, "G"{name=G}] & \C\arrow[l, shift left=.5ex, "F"{name=F}]
            \arrow[phantom, from=F, to=G, , "\scriptscriptstyle\boldsymbol{\top}"].
        \end{tikzcd}
    \]

\end{definition} 

\begin{definition}
The {\bf nerve} of a category ${\cal C}$ is the set of composable morphisms of length $n$, for $n \geq 1$.  Let $N_n({\cal C})$ denote the set of sequences of composable morphisms of length $n$.  

\[ \{ C_o \xrightarrow[]{f_1} C_1 \xrightarrow[]{f_2} \ldots \xrightarrow[]{f_n} C_n \ | \ C_i \ \mbox{is an object in} \ {\cal C}, f_i \ \mbox{is a morphism in} \ {\cal C} \} \] 
\end{definition}

The set of $n$-tuples of composable arrows in {\cal C}, denoted by $N_n({\cal C})$,  can be viewed as a functor from the simplicial object $[n]$ to ${\cal C}$.  Note that any nondecreasing map $\alpha: [m] \rightarrow [n]$ determines a map of sets $N_m({\cal C}) \rightarrow N_n({\cal C})$.  The nerve of a category {\cal C} is the simplicial set $N_\bullet: \Delta \rightarrow N_n({\cal C})$, which maps the ordinal number object $[n]$ to the set $N_n({\cal C})$.

The importance of the nerve of a category comes from a key result \cite{kerodon}, showing it defines a full and faithful embedding of a category: 

\begin{theorem}
\cite[\href{https://kerodon.net/tag/002Y}{Tag 002Y}]{kerodon}: The {\bf nerve functor} $N_\bullet: {\bf Cat} \rightarrow {\bf Set}$ is fully faithful. More specifically, there is a bijection $\theta$ defined as: 

\[ \theta: {\bf Cat}({\cal C}, {\cal C'}) \rightarrow {\bf Set}_\Delta (N_\bullet({\cal C}), N_\bullet({\cal C'}) \] 
\end{theorem}

 In general, the functor $G$ from a simplicial object $X$ to a category ${\cal C}$ can be lossy. For example, we can define the objects of ${\cal C}$ to be the elements of $X_0$, and the morphisms of ${\cal C}$ as the elements $f \in X_1$, where $f: a \rightarrow b$, and $d_0 f = a$, and $d_1 f = b$, and $s_0 a, a \in X$ as defining the identity morphisms ${\bf 1}_a$. Composition in this case can be defined as the free algebra defined over elements of $X_1$, subject to the constraints given by elements of $X_2$. For example, if $x \in X_2$, we can impose the requirement that $d_1 x = d_0 x \circ d_2 x$. Such a definition of the left adjoint would be quite lossy because it only preserves the structure of the simplicial object $X$ up to the $2$-simplices. The right adjoint from a category to its associated simplicial object, in contrast, constructs a full and faithful embedding of a category into a simplicial set.  In particular, the  nerve of a category is such a right adjoint. 

\begin{example} 
Given a category ${\cal C}$, and an $n$-simplex $\sigma$ of the simplicial set $N_n({\cal C})$,  which we can identify with the sequence: 

\[ \sigma = C_o \xrightarrow[]{f_1} C_1 \xrightarrow[]{f_2} \ldots \xrightarrow[]{f_n} C_n \] 

the face operator $d_0$ applied to $\sigma$ yields the sequence 

\[ d_0 \sigma = C_1 \xrightarrow[]{f_2} C_2 \xrightarrow[]{f_3} \ldots \xrightarrow[]{f_n} C_n \] 

where the object $C_0$ is ``deleted" along with the morphism $f_0$ leaving it. 

\end{example} 

\begin{example} 
Given a category ${\cal C}$, and an $n$-simplex $\sigma$ of the simplicial set $N_n({\cal C})$,  which we can identify with the sequence: 

\[ \sigma = C_o \xrightarrow[]{f_1} C_1 \xrightarrow[]{f_2} \ldots \xrightarrow[]{f_n} C_n \] 

the face operator $d_n$ applied to $\sigma$ yields the sequence 

\[ d_n \sigma = C_0 \xrightarrow[]{f_1} C_1 \xrightarrow[]{f_2} \ldots \xrightarrow[]{f_{n-1}} C_{n-1} \] 

where the object $C_n$ is ``deleted" along with the morphism $f_n$ entering it. 

\end{example} 

\begin{example} 
Given a category  ${\cal C}$, and an $n$-simplex $\sigma$ of the simplicial set $N_n({\cal C})$,  which we can identify with the sequence: 

\[ \sigma = C_o \xrightarrow[]{f_1} C_1 \xrightarrow[]{f_2} \ldots \xrightarrow[]{f_n} C_n \] 

the face operator $d_i, 0 < i < n$ applied to $\sigma$ yields the sequence 

\[ d_i \sigma = C_0 \xrightarrow[]{f_1} C_1 \xrightarrow[]{f_2} \ldots C_{i-1} \xrightarrow[]{f_{i+1} \circ f_i} C_{i+1} \ldots \xrightarrow[]{f_{n}} C_{n} \] 

where the object $C_i$ is ``deleted" and the morphisms $f_i$ is composed with morphism $f_{i+1}$.  

\end{example} 

\begin{example} 
Given a category ${\cal C}$, and an $n$-simplex $\sigma$ of the simplicial set $N_n({\cal C})$,  which we can identify with the sequence: 

\[ \sigma = C_o \xrightarrow[]{f_1} C_1 \xrightarrow[]{f_2} \ldots \xrightarrow[]{f_n} C_n \] 

the degeneracy operator $s_i, 0 \leq i \leq n$ applied to $\sigma$ yields the sequence 

\[ s_i \sigma = C_0 \xrightarrow[]{f_1} C_1 \xrightarrow[]{f_2} \ldots C_{i} \xrightarrow[]{{\bf 1}_{C_i}} C_{i} \xrightarrow[]{f_{i+1}} C_{i+1}\ldots \xrightarrow[]{f_{n}} C_{n} \] 

where the object $C_i$ is ``repeated" by inserting its identity morphism ${\bf 1}_{C_i}$. 

\end{example} 

\begin{definition} 
Given a category ${\cal C}$, and an $n$-simplex $\sigma$ of the simplicial set $N_n({\cal C})$,  which we can identify with the sequence: 

\[ \sigma = C_o \xrightarrow[]{f_1} C_1 \xrightarrow[]{f_2} \ldots \xrightarrow[]{f_n} C_n \] 

we define $\sigma$ as a {\bf degenerate} simplex if some $f_i$ in the above sequence is an identity morphism, in which case $C_i$ and $C_{i+1}$ are equal. 
\end{definition} 

\section{Horns of Simplicial Objects} 

One of the fundamental contributions of this paper is to show that structure discovery in causal inference and RL both involve solving extensions problems defined by commutative diagrams, which are defined as ``horn filling" in higher-order category theory \citep{weakkan,quasicats,kerodon}. 

 \subsection{Simplicial Subsets and Horns}
 
 We now describe ways of modeling subsets of simplicial objects, which are simplicial objects in their own right.  We introduce the crucial concept of the {\em boundary} and {\em horn} of a simplicial set. 
 
 \begin{definition}
 The {\bf standard simplex} $\Delta^n$ is the simplicial set defined by the construction 
 
 \[ ([m] \in \Delta) \mapsto {\bf Hom}_\Delta([m], [n]) \] 
 
 By convention, $\Delta^{-1} \coloneqq \emptyset$. The standard $0$-simplex $\Delta^0$ maps each $[n] \in \Delta^{op}$ to the single element set $\{ \bullet \}$. 
 \end{definition}
 
 \begin{definition}
 Let $S_\bullet$ denote a simplicial set. If for every integer $n \geq 0$, we are given a subset $T_n \subseteq S_n$, such that the face and degeneracy maps 
 
 \[ d_i: S_n \rightarrow S_{n-1} \ \ \ \ s_i: S_n \rightarrow S_{n+1} \] 
 
 applied to $T_n$ result in 
 
 \[ d_i: T_n \rightarrow T_{n-1} \ \ \ \ s_i: T_n \rightarrow T_{n+1} \] 
 
 then the collection $\{ T_n \}_{n \geq 0}$ defines a {\bf simplicial subset} $T_\bullet \subseteq S_\bullet$
 \end{definition}
 
 \begin{definition}
 The {\bf boundary} is a simplicial set $(\partial \Delta^n): \Delta^{op} \rightarrow {\bf Set}$ defined as
 
 \[ (\partial \Delta^n)([m]) = \{ \alpha \in {\bf Hom}_\Delta([m], [n]): \alpha \ \mbox{is not surjective} \} \]
 \end{definition}
 
 Note that the boundary $\partial \Delta^n$ is a simplicial subset of the standard $n$-simplex $\Delta^n$. 
 
 \begin{definition}
 The {\bf Horn} $\Lambda^n_i: \Delta^{op} \rightarrow {\bf Set}$ is defined as
 
 \[ (\Lambda^n_i)([m]) = \{ \alpha \in {\bf Hom}_\Delta([m],[n]): [n] \not \subseteq \alpha([m]) \cup \{i \} \} \] 
 \end{definition}
 
 Intuitively, the Horn $\Lambda^n_i$ can be viewed as the simplicial subset that results from removing the interior of the $n$-simplex $\Delta^n$ together with the face opposite its $i$th vertex.   Lifting problems provide elegant ways to define basic notions in a wide variety of areas in mathematics. For example, the notion of injective and surjective functions, the notion of separation in topology, and many other basic constructs can be formulated as solutions to lifting problems. \\
 
 Let us illustrate this abstract definition with the following diagrams. Consider the problem of composing $1$-dimensional simplices  to form a $2$-dimensional simplicial object. Each simplicial subset of an $n$-simplex induces a  a {\em horn} $\Lambda^n_k$, where  $ 0 \leq k \leq n$. Intuitively, a horn is a subset of a simplicial object that results from removing the interior of the $n$-simplex and the face opposite the $i$th vertex. Consider the three horns defined below. The dashed arrow  $\dashrightarrow$ indicates edges of the $2$-simplex $\Delta^2$ not contained in the horns. 

\begin{center}
 \begin{tikzcd}[column sep=small]
& \{0\}  \arrow[dl] \arrow[dr] & \\
  \{1 \} \arrow[rr, dashed] &                         & \{ 2 \} 
\end{tikzcd} \hskip 0.5 in 
 \begin{tikzcd}[column sep=small]
& \{0\}  \arrow[dl] \arrow[dr, dashed] & \\
  \{1 \} \arrow{rr} &                         & \{ 2 \} 
\end{tikzcd} \hskip 0.5in 
 \begin{tikzcd}[column sep=small]
& \{0\}  \arrow[dl, dashed] \arrow[dr] & \\
  \{1 \} \arrow{rr} &                         & \{ 2 \} 
\end{tikzcd}
\end{center}

The inner horn $\Lambda^2_1$ is the middle diagram above, and admits an easy solution to the ``horn filling" problem of composing the simplicial subsets. The two outer horns on either end pose a more difficult challenge. A considerable elaboration of the theoretical machinery in category theory is required to describe the various solutions proposed, which led to different ways of defining higher-order category theory \citep{weakkan,quasicats,kerodon}, which we summarize below. 
 
 \begin{definition}
 Let ${\cal C}$ be a category. A {\bf lifting problem} in ${\cal C}$ is a commutative diagram $\sigma$ in ${\cal C}$. 
 
 \begin{center}
 \begin{tikzcd}
  A \arrow{d}{f} \arrow{r}{\mu}
    & X \arrow[red]{d}{p} \\
  B  \arrow[red]{r}[blue]{\nu}
&Y \end{tikzcd}
 \end{center} 
 \end{definition}
 
 \begin{definition}
 Let ${\cal C}$ be a category. A {\bf solution to a lifting problem} in ${\cal C}$ is a morphism $h: B \rightarrow X$ in ${\cal C}$ satisfying $p \circ h = \nu$ and $h \circ f = \mu$ as indicated in the diagram below.
 
 \begin{center}
 \begin{tikzcd}
  A \arrow{d}{f} \arrow{r}{\mu}
    & X \arrow[red]{d}{p} \\
  B \arrow[ur,dashed, "h"] \arrow[red]{r}[blue]{\nu}
&Y \end{tikzcd}
 \end{center} 
 \end{definition}
 
 \begin{definition}
 Let ${\cal C}$ be a category. If we are given two morphisms $f: A \rightarrow B$ and $p: X \rightarrow Y$ in ${\cal C}$, we say that $f$ has the {\bf left lifting property} with respect to $p$, or that p has the {\bf right lifting property} with respect to f if for every pair of morphisms $\mu: A \rightarrow X$ and $\nu: B \rightarrow Y$ satisfying the equations $p \circ \mu = \nu \circ f$, the associated lifting problem indicated in the diagram below.
 
 \begin{center}
 \begin{tikzcd}
  A \arrow{d}{f} \arrow{r}{\mu}
    & X \arrow[red]{d}{p} \\
  B \arrow[ur,dashed, "h"] \arrow[red]{r}[blue]{\nu}
&Y \end{tikzcd}
 \end{center} 
 
 admits a solution given by the map $h: B \rightarrow X$ satisfying $p \circ h = \nu$ and $h \circ f = \mu$. 
 \end{definition}
 
 \begin{example}
 Given the paradigmatic non-surjective morphism $f: \emptyset \rightarrow \{ \bullet \}$, any morphism p that has the right lifting property with respect to f is a {\bf surjective mapping}. 
 
 \begin{center}
 \begin{tikzcd}
  \emptyset \arrow{d}{f} \arrow{r}{\mu}
    & X \arrow[red]{d}{p} \\
  \{ \bullet \} \arrow[ur,dashed, "h"] \arrow[red]{r}[blue]{\nu}
&Y \end{tikzcd}
 \end{center} 

 \end{example}
 
 \begin{example}
 Given the paradigmatic non-injective morphism $f: \{ \bullet, \bullet \} \rightarrow \{ \bullet \}$, any morphism p that has the right lifting property with respect to f is an {\bf injective mapping}. 
 \begin{center}
 \begin{tikzcd}
  \{\bullet, \bullet \} \arrow{d}{f} \arrow{r}{\mu}
    & X \arrow[red]{d}{p} \\
  \{ \bullet \} \arrow[ur,dashed, "h"]  \arrow[red]{r}[blue]{\nu}
&Y \end{tikzcd}
 \end{center} 

 \end{example}
 
\begin{definition}
 If $S$ is a collection of morphisms in category  ${\cal C}$, a morphism $f: A \rightarrow B$ has the {\bf left lifting property with respect to S} if it has the left lifting property with respect to every morphism in $S$. Analogously, we say a morphism $p: X \rightarrow Y$ has the {\bf right lifting property with respect to S} if it has the right lifting property with respect to every morphism in $S$.
 \end{definition}
 
 \begin{definition}
 Let $C$ and $C'$ be a pair of objects in a category ${\cal C}$. We say $C$ is {\bf a retract} of $C'$ if there exists maps $i: C \rightarrow C'$ and $r: C' \rightarrow C$ such that $r \circ i = \mbox{id}_{\cal C}$. 
 \end{definition}
 
 \begin{definition}
 Let ${\cal C}$ be a category. We say a morphism $f: C \rightarrow D$ is a {\bf retract of another morphism} $f': C \rightarrow D$ if it is a retract of $f'$ when viewed as an object of the functor category ${\bf Hom}([1], {\cal C})$. A collection of morphisms $T$ of ${\cal C}$ is {\bf closed under retracts} if for every pair of morphisms $f, f'$ of ${\cal C}$, if $f$ is a retract of $f'$, and $f'$  is in $T$, then $f$ is also in $T$. 
 \end{definition}
 
 \subsection{Fibrations and Kan Complexes} 
 
 \begin{definition}
 Let $f: X \rightarrow S$ be a morphism of simplicial sets. We say $f$ is a {\bf Kan fibration} if, for each $n > 0$, and each $0 \leq i \leq n$, every lifting problem 
 
 \begin{center}
 \begin{tikzcd}
  \Lambda^n_i \arrow{d}{} \arrow{r}{\sigma_0}
    & X \arrow[red]{d}{f} \\
  \Delta^n \arrow[ur,dashed, "\sigma"] \arrow[red]{r}[blue]{\bar{\sigma}}
&S \end{tikzcd}
 \end{center} 
 
 admits a solution. More precisely, for every map of simplicial sets $\sigma_0: \Lambda^n_i \rightarrow X$ and every $n$-simplex $\bar{\sigma}: \Delta^n \rightarrow S$ extending $f \circ \sigma_0$, we can extend $\sigma_0$ to an $n$-simplex $\sigma: \Delta^n \rightarrow X$ satisfying $f \circ \sigma = \bar{\sigma}$. 
 \end{definition}
 
 \begin{example}
Given a simplicial set $X$, then a projection map $X \rightarrow \Delta^0$ that is a Kan fibration is called a {\bf Kan complex}. 
\end{example} 

\begin{example}
Any isomorphism between simplicial sets is a Kan fibration. 
\end{example}

\begin{example}
The collection of Kan fibrations is closed under retracts. 
\end{example}

\begin{definition}
  Let X and Y be simplicial sets, and suppose we are given a pair of morphisms $f_0, f_1: X \rightarrow Y$. A {\bf homotopy} from $f_0$ to $f_1$ is a morphism $h: \Delta^1 \times X \rightarrow Y$ satisfying $f_0 = h |_{{0} \times X}$ and $f_1 = h_{ 1 \times X}$. 
 \end{definition}

\subsection{Higher-order Categories} 

We now formally introduce higher-order categories, building on the framework proposed in a number of formalisms \citep{weakkan,quasicats,kerodon}. 

\begin{definition}\citep{kerodon}
\label{ic} 
An $\infty$-category is a simplicial object $S_\bullet$ which satisfies the following condition: 

\begin{itemize} 
\item For $0 < i < n$, every map of simplicial sets $\sigma_0: \Lambda^n_i \rightarrow S_\bullet$ can be extended to a map $\sigma: \Delta^n \rightarrow S_\bullet$. 
\end{itemize} 
\end{definition}

This definition emerges out of a common generalization of two other conditions on a simplicial set $S_i$: 

\begin{enumerate} 
\item {\bf Property K}: For $n > 0$ and $0 \leq i \leq n$, every map of simplicial sets $\sigma_0: \Lambda^n_i \rightarrow S_\bullet$ can be extended to a map $\sigma: \Delta^n \rightarrow S_\bullet$. 

\item {\bf Property C}:  for $0 < 1 < n$, every map of simplicial sets $\sigma_0: \Lambda^n_i \rightarrow S_i$ can be extended {\em uniquely} to a map $\sigma: \Delta^n \rightarrow S_\bullet$. 
\end{enumerate} 

Simplicial objects that satisfy property K were defined above to be Kan complexes. Simplicial objects that satisfy property C above can be identified with the nerve of a category, which yields a full and faithful embedding of a category in the category of sets.  Definition~\ref{ic} generalizes both of these definitions, and was called a {\em quasicategory} in \citep{quasicats} and {\em weak Kan complexes} in \citep{weakkan} when ${\cal C}$ is a category.

\section{Universal Reinforcement Learning}

In this section, we briefly review the framework of universal decision models \citep{sm:udm}, where a decision making object is defined as a particular type of category.  We will focus in particular on predictive state representations (PSRs) \cite{singh-uai04}, but we will include a brief review of Markov decision processes (MDPs), which are a stochastic dynamical system widely used in the literature on reinforcement learning. We show that predictive state representations define a simplicial object representation of a category, and prove the Universal PSR theorem based on the fact that the nerve of a PSR defines a full and faithful embedding of the category of PSRs in the category of sets. We show that simplicial object representations of PSRs define a quasicategory. 

\subsection{ UDMs based on Markov Decision Processes}

We now briefly describe the category of UDMs, where each object represents a (finite) Markov decision process (MDP) \citep{DBLP:books/wi/Puterman94}.  Recall that an MDP is defined by a tuple $\langle S, A, \Psi, P, R \rangle$, where $S$ is a discrete set of states, $A$ is the discrete set of actions, $\Psi \subset S \times A$ is the set of admissible state-action pairs, $P: \Psi \times S \rightarrow [0,1]$ is the transition probability function specifying the one-step dynamics of the model, where $P(s,a,s')$ is the transition probability of moving from state $s$ to state $s'$ in one step under action $a$, and $R: \Psi \rightarrow \mathbb{R}$ is the expected reward function, where $R(s,a)$ is the expected reward for executing action $a$ in state $s$. MDP homomorphisms can be viewed as a principled way of abstracting the state (action) set of an MDP into a ``simpler" MDP that nonetheless preserves some important properties, usually referred to as the stochastic substitution property (SSP). 

\begin{definition}
A UDM MDP homomorphism \citep{DBLP:conf/ijcai/RavindranB03} from object  $M = \langle S, A, \Psi, P, R \rangle$ to $M' = \langle S', A', \Psi', P', R' \rangle$, denoted $h: M \twoheadrightarrow M'$, is defined by a tuple of surjections $\langle f, \{g_s | s \in S \} \rangle$, where $f: S \twoheadrightarrow S', g_s: A_s \twoheadrightarrow A'_{f(s)}$, where $h((s,a)) = \langle f(s), g_s(a) \rangle$, for $s \in S$, such that the stochastic substitution property and reward respecting properties below are respected: 
\begin{eqnarray} 
\label{mdp-hom}
P'(f(s), g_s(a), f(s')) = \sum_{s" \in [s']_f} P(s, a, s") \\
R'(f(s), g_s(a)) = R(s, a)
\end{eqnarray} 
\end{definition}

Given this definition, the following result is straightforward to prove. 

\begin{theorem}
The UDM category ${\cal C}_{\mbox{MDP}}$ is defined as one where each object $c$ is defined by an MDP, and morphisms are given by MDP homomorphisms defined by Equation~\ref{mdp-hom}. 
\end{theorem}

{\bf Proof:} Note that the composition of two MDP homomorphisms $h: M_1 \rightarrow M_2$ and $h': M_2 \rightarrow M_3$ is once again an MDP homomorphism $h' \circ h: M_1 \rightarrow M_3$. The identity homomorphism is easy to define, and MDP homomorphisms, being surjective mappings, obey associative properties. $\qed$

\subsection{UDM Category of Predictive State Representations} 

\begin{figure}[t] 
\centering
\vskip 0.1in
\begin{minipage}{0.9\textwidth}
\includegraphics[scale=0.4]{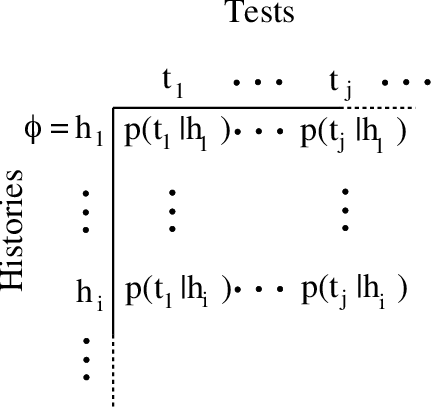}
\includegraphics[scale=0.35]{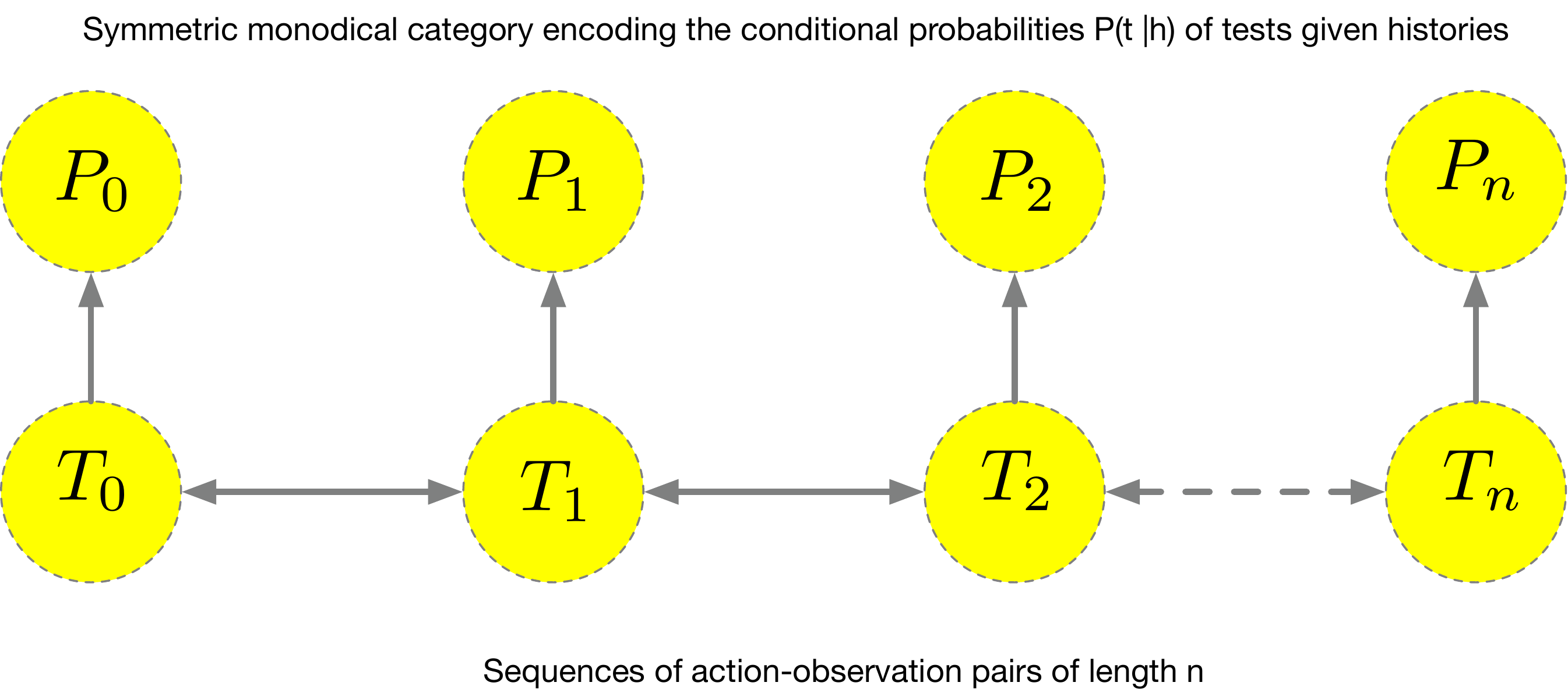}
\end{minipage}
\caption{Left: A predictive state representation (PSR) can be defined in terms of its underlying systems dynamics matrix \citep{singh-uai04}.  The columns of this (infinite) matrix are tests $t = a_1 o_1 \ldots a_n o_n$ and  the rows are histories $h = a_1 o_1 \ldots a_k o_k$. Right: the simplicial object encoded by a PSR. $T_n$ represents the $n$-simplicial object of tests or histories, and $P_n$ represents a symmetric monoidal category \cite{fong2018seven} that can be used to encode conditional probabilities $P(t | h)$ of each test $t$ given its history $h$. Note that as in Figure~\ref{simpobj}, there are $n$ arrows leading from the $n$-simplex $T_n$ back to the $n-1$ simplex $T_{n-1}$ as well as to the $n+1$-simplex $T_{n+1}$, which are not shown for clarity. The conditional probability for each test $t$ as a function of history $h$, $P(t | h)$ can be defined using an enriched symmetric monoidal category, exactly as defined in \citep{Bradley_2022}, which defines sentence completions in English by deep NLP networks using conditional probabilities as enriched categories. \label{psr-sdm}  }
 \end{figure}

We now define the UDM (sub)category ${\cal C}_{\mbox{PSR}}$ of predictive state representations \citep{DBLP:journals/jmlr/ThonJ15}, based on the notion of homomorphism defined for PSRs proposed in  \citep{DBLP:conf/aaai/SoniS07}.  Recall that a PSR is (in the simplest case) a discrete controlled dynamical system, characterized by a finite set of actions $A$, and observations $O$. At each clock tick $t$, the agent takes an action $a_t$ and receives an observation $o_t \in O$. A {\em history} is defined as a sequence of actions and observations $h = a_1 o_1 \ldots a_k o_k$. A {\em test} is a possible sequence of future actions and observations $t = a_1 o_1 \ldots a_n o_n$. A test is successful if the observations $o_1 \ldots o_n$ are observed in that order, upon execution of actions $a_1 \ldots a_n$. The probability $P(t | h)$ is a prediction of that a test $t$ will succeed from history $h$. 

A state $\psi$ in a PSR is a vector of predictions of a suite of {\em core tests} $\{q_1, \ldots, q_k \}$. The prediction vector $\psi_h = \langle P(q_1 | h) \ldots P(q_k | h) \rangle$ is a sufficient statistic, in that it can be used to make predictions for any test. More precisely, for every test $t$, there is a $1 \times k$ projection vector $m_t$ such that $P(t | h) = \psi_h . m_t$ for all histories $h$. The entire predictive state of a PSR can be denoted $\Psi$. 

\begin{definition}
\label{psr-homo}
In the UDM category ${\cal C}_{\mbox{PSR}}$ defined by PSR objects, the morphism from object $\Psi$ to another  $\Psi'$ is defined by a tuple of surjections $\langle f, v_\psi(a)\rangle$, where $f: \Psi \rightarrow \Psi'$ and $v_\psi: A \rightarrow A'$ for all prediction vectors $\psi \in \Psi$ such that 
\begin{equation}
    P(\psi' | f(\psi), v_\psi(a)) = P(f^{-1}(\psi') | \psi, a) 
\end{equation}
for all $\psi' \in \Psi, \psi \in \Psi, a \in A$. 
\end{definition}

\begin{theorem}
\label{psr}
The UDM category ${\cal C}_{\mbox{PSR}}$ is defined by making each object $c$ represent a PSR, where the morphisms between two PSRs $h: c \rightarrow d$ is defined by the PSR homomorphism defined in \citep{DBLP:conf/aaai/SoniS07}. 
\end{theorem}

 {\bf Proof:} Once again, given the homomorphism definition in Definition~\ref{psr-homo}, the UDM category ${\cal P}_{\mbox{PSR}}$ is easy to define, given the surjectivity of the associated mappings $f$ and $v_\psi$. $\qed$
 
 \subsection{PSR Test Discovery as Horn Filling of a Simplicial Object}

The extensive literature on PSRs has explored ways of constructing the set of core tests needed to serve as a sufficient statistic for a full and faithful embedding of any dynamical system that can be represented as a PSR. Here, we just want to illustrate that the process of elaborating a partial test $t = a_1 o_1 \ldots a_i o_i$ by adding a new candidate action observation pair $a_{i+1} o_{i+1}$ can be viewed as filling the horn of a simplicial object. 

\begin{definition}
 A {\bf predictive horn} of a PSR $\Lambda^n_i: \Delta^{op} \rightarrow {\cal C}_{PSR}$ is defined as
 as the simplicial subset of the functor mapping $[n]$ to the category ${\cal C}_{PSR}$ defining a PSR model. Recall the horn removes the interior of an $n$-simplex $\Delta^n$ together with the face opposite its $i$th vertex.  
 \end{definition}

Let us illustrate this notion of horn filling using the same diagram as shown earlier, but this time, interpreting the morphisms shown below as tests in a PSR. 

\begin{center}
 \begin{tikzcd}[column sep=small]
& \{0\}  \arrow[dl] \arrow[dr] & \\
  \{1 \} \arrow[rr, dashed] &                         & \{ 2 \} 
\end{tikzcd} \hskip 0.5 in 
 \begin{tikzcd}[column sep=small]
& \{0\}  \arrow[dl] \arrow[dr, dashed] & \\
  \{1 \} \arrow{rr} &                         & \{ 2 \} 
\end{tikzcd} \hskip 0.5in 
 \begin{tikzcd}[column sep=small]
& \{0\}  \arrow[dl, dashed] \arrow[dr] & \\
  \{1 \} \arrow{rr} &                         & \{ 2 \} 
\end{tikzcd}
\end{center}

The inner horn $\Lambda^2_1$ is the middle diagram above, and admits an easy solution to the ``horn filling" problem of composing the simplicial subsets. The two outer horns on either end pose a more difficult challenge. Stated in the language of PSRs, the inner horn filling problem represents the problem of composing two tests, defined as images of abstract tests $t_{01}: 0 \rightarrow 1$ and $t_{12}: 1 \rightarrow 2$ into the category defined by a PSR. Each image  of an abstract test leads to an actual test in a PSR model, for example $F(t_{01}) = a_i o_i$ and $F(t_{12}) = a_j o_j$. These two tests can be composed to form a test $F(t_{01} \circ t_{12}) = F(t_{02}) = a_i o_i a_j o_j$. The outer horn filling problem, in contrast, involves inducing a missing test that ``fills in" between a test of length $2$ and a test of length $1$. 

First, let us define more precisely how to construct the associated simplicial object from a PSR. We can define this construction following the systems dynamics matrix in Figure~\ref{psr-sdm}. Both tests and histories are viewed as $n$-simplices, constructed from sequences of action-observation pairs. For example, in Figure~\ref{psr-sdm}, the $1$-simplex encodes a unit test or history $a_i o_j$ of a single action observation pair. The conditional probabilities $P(t | h)$ of each test $t$ given its history $h$ is defined using an enriched symmetric monoidal category $P$. Basically, we are treating the process of combining a test $t$ with a history $h$ as a morphism from $h$ to $t \circ h$. The set of all morphisms ${\bf Hom}_{\cal C}(h,th)$ is defined as an enriched symmetric monoidal category that can encode conditional probabilities. \citep{Bradley_2022} gives a detailed treatment of how to model conditional probabilities for the problem of defining the compositional structure of deep NLP models, where for each fragment of an English sentence, like $h=$ ``I went to the grocery store...", each possible completion, such as $t =$ ``to get some milk" is defined as a conditional probability $P(t|h)$. This problem is exactly the same as defining the conditional probability of PSR tests, and we refer the reader to \citep{Bradley_2022} for additional details.

\begin{definition}
The {\bf nerve} of a PSR ${\cal C}$ is the set of sequences of composable tests of length $n$, for $n \geq 1$.  Let $N_n({\cal C})$ denote the set of sequences of composable tests 

\[ \{ C_o \xrightarrow[]{f_1} C_1 \xrightarrow[]{f_2} \ldots \xrightarrow[]{f_n} C_n \ | \ C_i \ \mbox{is an object in} \ {\cal C}, f_i \ \mbox{is a morphism in} \ {\cal C} \} \] 

\[
        \begin{tikzcd}
            {\cal N}_{PSR} ({\cal C}) \arrow[r, shift left=1ex, "G"{name=G}] & \C_{PSR}\arrow[l, shift left=.5ex, "F"{name=F}]
            \arrow[phantom, from=F, to=G, , "\scriptscriptstyle\boldsymbol{\top}"].
        \end{tikzcd}
    \]
\end{definition}

We can use this result to abstractly define a universal PSR theorem, which states that any PSR can be defined completely by its nerve. 

\begin{theorem} {\bf Universal PSR Theorem:}
The {\bf nerve functor} defined by a PSR $N_\bullet: {\bf Cat}_{PSR} \rightarrow {\bf Set}$ is fully faithful. More specifically, there is a bijection $\theta$ defined as: 

\[ \theta: {\bf Cat}({\cal C}_{PSR}, {\cal C'}_{PSR}) \rightarrow {\bf Set}_\Delta (N_\bullet({\cal C}_{PSR}), N_\bullet({\cal C'}_{PSR}) \] 
\end{theorem}

\begin{theorem} {\bf PSRs form a quasicategory:}
The {\bf nerve functor} defined by a PSR $N_\bullet: {\bf Cat}_{PSR} \rightarrow {\bf Set}$ forms a quasicategory \cite{quasicats}. 
\end{theorem}

{\bf Proof:} The proof of both theorems follows directly from more basic results establishing that the nerve of a category is a fully faithful embedding of it. Since we showed above in Theorem~\ref{psr} that PSRs form a category, the nerve embedding is fully faithful. It also follows from \citep{quasicats} that the nerve functor defined by the category of PSRs defines a quasicategory. $\bullet$

In particular, the nerve of a PSR defines a quasicategory, and this result shows that its nerve is a full and faithful embedding of a category as a simplicial object. The significance of this theorem is it shows that a considerable amount of theoretical machinery in higher-order category theory can be brought to bear on the problem of structure discovery of a PSR. There are many elaborations of this basic result, which we will postpone to a subsequent paper.  The definition of nerve leads to an important topological characterization of any category. 

\begin{definition}
The {\bf classifying space} of a PSR ${\cal C}$ is the topological space defined by its nerve functor $|N_\bullet({\cal C})|$. 
\end{definition}

The classifying space of a category gives a way to define an algebraic invariant of a PSR, which we define below as its singular homology. 

\subsection{Singular Homology of a PSR} 

We will now describe the singular homology of a PSR.  First, we need to define more concretely the topological $n$-simplex that provides a concrete way to attach a topology to a simplicial object. Our definitions below build on those given in \citep{kerodon}. For each integer $n$, define the topological space $|\Delta_n|$ realized by the object $\Delta_n$ as 

\[ |\Delta_n| = \{t_0, t_1, \ldots, t_n  \in \mathbb{R}^{n+1}: t_0 + t_1 + \ldots + t_n = 1 \} \] 

This is the familiar $n$-dimensional simplex over $n$ variables. For any PSR model, its classifying space $|{\cal N}_\bullet({\cal C})|$ defines a topological space. We can now define the {\em singular} $n$-simplex is a continuous mapping $\sigma: |\Delta_N| \rightarrow |{\cal N}_\bullet({\cal C})|$. Every singular $n$-simplex $\sigma$ induces a collection of $n-1$-dimensional simplices called {\em faces}, denoted as 

\[ d_i \sigma(t_0, \ldots, t_{n-1}) = (t_0, t_1, \ldots, t_{i-1}, 0, t_i, \ldots, t_{n-1}) \] 

Define the set of all morphisms $\mbox{Sing}_n(X) = {\bf Hom}_{\bf Top}(\Delta_n, |{\cal N}_\bullet({\cal C})|)$ as the set of singular $n$-simplices of $|{\cal N}_\bullet({\cal C})|$. 

\begin{definition}
For any topological space defined by a PSR $|{\cal N}_\bullet({\cal C})|$,  the {\bf singular homology groups} $H_*(|{\cal N}_\bullet({\cal C})|; {\bf Z})$ are defined as the homology groups of a chain complex 

\[ \ldots \xrightarrow[]{\partial} {\bf Z}(\mbox{Sing}_2(|{\cal N}_\bullet({\cal C})|)) \xrightarrow[]{\partial} {\bf Z}(\mbox{Sing}_1(|{\cal N}_\bullet({\cal C})|)) \xrightarrow[]{\partial} {\bf Z}(\mbox{Sing}_0(|{\cal N}_\bullet({\cal C})|)) \] 

where ${\bf Z}(\mbox{Sing}_n(|{\cal N}_\bullet({\cal C})|))$ denotes the free Abelian group generated by the set $\mbox{Sing}_n(|{\cal N}_\bullet({\cal C})|)$ and the differential $\partial$ is defined on the generators by the formula 

\[ \partial (\sigma) = \sum_{i=0}^n (-1)^i d_i \sigma \] 
\end{definition}

Intuitively, a chain complex builds a sequence of vector spaces that can be used to construct an algebraic invariant of a PSR from its classifying space  by choosing the left {\bf k} module ${\bf Z}$ to be a vector space. Each differential $\partial$ then becomes a linear transformation whose representation is constructed by modeling its effect on the basis elements in each ${\bf Z}(\mbox{Sing}_n(X))$. 

\section{Universal Causal Model}

We now state more formally the Universal Causal Model (UCM) framework \citep{sm:uc}, which provides some background to  simplicial objects formulation below.  

\begin{figure}[t]
\centering
\begin{minipage}{0.7\textwidth}
\centering
\hfill
\includegraphics[scale=0.45]{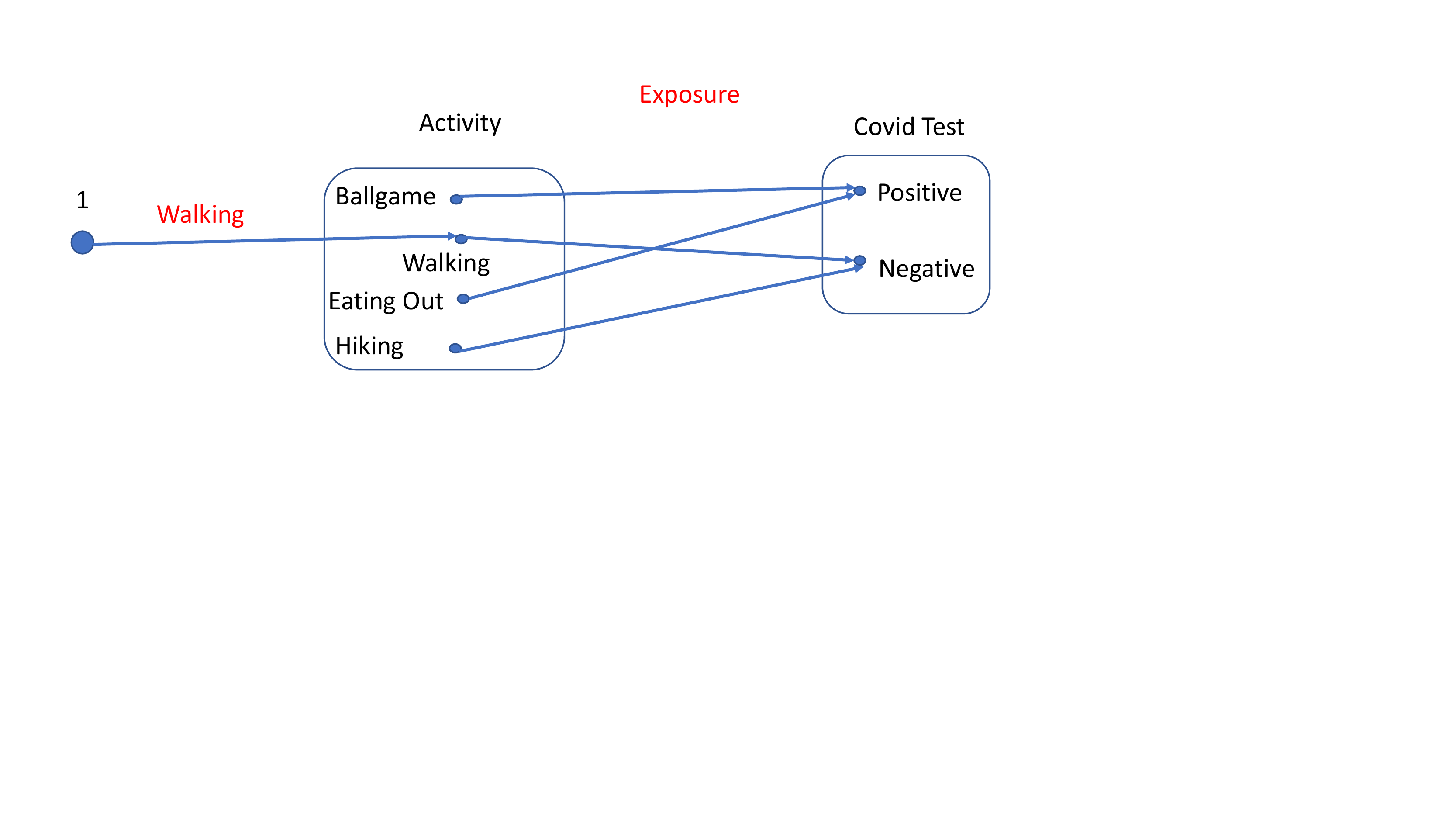}
\end{minipage} \vskip -1.5in
\caption{A simple category ${\cal C}$ for studying causal effects of exposure to getting Covid-19 infections. The morphism {\bf Exposure } maps a finite set object {\bf Activity} to the (Boolean) object  {\bf Covid Test}, where the morphism could represent any arbitrary  measure-theoretic \cite{witsenhausen:1975, heymann:if}, probabilistic \cite{pearl-book}, or topological relationship \cite{sm:homotopy}.  In category theory, an ``element" of a set (such as {\bf Walking}) is formally denoted by a morphism from the unit object {\bf 1} (which has one object and one identity morphism) to the set that is defined by the label it maps to. Causal  interventions are viewed as morphisms into an object, such as determining the causal effect of {\bf Activity} on {\bf Covid Test} by intervening on the {\bf Activity} variable by setting it to {\bf Walking}. In this paper, we model interventions as elementary face and degeneracy mappings on  simplicial objects.}  
\label{covid-category}
\end{figure}

\begin{definition}
A {\bf universal causal model} (UCM) is defined as a tuple $\langle {\cal C}, {\cal X}, {\cal I}, {\cal O}, {\cal E} \rangle$ where each of the components is specified as follows: 

\begin{enumerate} 
\item ${\cal C}$ is a category of causal objects that interact with each other, and their patterns of interactions are captured by a set of morphisms {\bf Hom}$_{\cal C}(X,Y)$ between object $X$ and $Y$. Categories where the {\bf Hom} morphisms can be defined as a set are referred to as {\em locally small} categories, which are the ones of primary interest to us in this paper. In particular, a {\cal V}-enriched category ${\cal C}$ is one where the {\bf Hom} morphisms are defined over the category {\cal V}, which allows capturing additional structure that might exist in the set of morphisms. 

\item ${\cal X}$ is a set of {\em construction} objects that allow constructing complex causal models from elementary parts. Examples include monoidal tensor product of two categories ${\cal M} \otimes {\cal M}$, Galois extensions and profunctors between two partially ordered sets to represent resource models,  co-limits and decorated cospans to represent electric circuits, and so on. In this paper, we use the construction tools from the theory  of simplicial objects. 

\item ${\cal I}$ is an intervention category of intervention objects, which map from some category ${\cal E}$ of experimental designs into the causal category ${\cal C}$ to implement an experimental design. Interventions can be simple, such as setting the value of a variable to a specific value (see Figure~\ref{covid-category}), in which case ${\cal I}$ is a comma category, or they can be more complex, such as setting the value of a variable to some arbitrary marginal distribution, such as the edge-intervention model proposed by \citet{janzing}.  There is a large literature on treatment planning in causal inference, and many of the state of the art applications of causal inference, such as bipartite experiments \cite{zigler2018bipartite} use sophisticated experimental designs. Any of these experimental designs can be accommodated in our functorial intervention framework. In this paper, we use the elementary face and degeneracy operators defined earlier on simplicial objects. 

\item ${\cal O}$ is a functor category of observation objects, which provide a (partial, perhaps noisy) view of causal objects in ${\cal C}$. For example, an object in the covariant functor category of presheafs {\cal Hom}$_{\cal C}(X, -)$ of morphisms out of $X$ can be viewed as a measurement of the ``state" of $X$.  More general examples include {\em bisimulation morphisms} used in the literature in reinforcement learning and software systems \cite{sm:udm}. \citet{adams} propose modeling observation as an order-preserving function $\Phi$ from a system ${\cal S}$ modeled by a lattice to an observation structure ${\cal O}$, which is also a lattice. They define the composition of subsystems into a larger system as a {\em join} $S_1 \vee S_2$, where a system exhibits a generative effect if $\Phi(S_1 \vee S_2) \neq \Phi(S_1) \vee \Phi(S_2)$. 

\item ${\cal E}$ is a functor category of evaluation objects, which map a given causal model into an evaluation category for the purposes of evaluating the effects of an intervention. For an example, for a network economy \cite{nagurney:vibook}, the evaluation functor {\cal E} is defined as a mapping from the causal network economy model to the category of real-valued vector spaces defined by the vector field $F$ of a variational inequality (VI) $\langle F(x^*), (x - x^*) \rangle \geq 0, \forall x \in$. Here, the vector field $F: {\cal C} \rightarrow \mathbb{R}^n$ is a functor from the causal category ${\cal C}$ of network economy models (see \cite{sm:udm}) to the category of $n$-dimensional Euclidean space. Solving a VI means finding equilibrium flows $x^*$ on the network that represent stable patterns of trade. For example, due to the recent war in Ukraine, global supply chains from grain to natural gas and oil have been disrupted, and this intervention requires the global economic system to find a new equilibrium. The problem of studying causal interventions in network economies was recently studied by us in a previous paper \cite{sm:causal-network-econ}, and we refer the reader to that paper for more details. In a causal DAG model, the evaluation functor maps the causal category into the category of probability spaces (i.e., the  rules of Pearl's do-calculus \cite{pearl-book} give conditions under which $P(Y | do(X)) = P(Y | X)$).  
\end{enumerate} 

\end{definition}

\subsection{Causal Inference over Simplicial Objects} 

We now turn to formulating causal discovery in terms of adjoint functors between a category, defining a universal causal model \citep{sm:uc}, and a simplicial object. We define causal interventions as   images of  the elementary injective maps $d_i: [n] \rightarrow [n+1]$, which skips element $i$ in its image, and surjective maps, $s_j: [n] \rightarrow [n-1]$, which repeats element $i$, in the category of ordinal numbers. Defining causal inference over simplicial objects requires choosing a compositional  structure over higher-order simplicial objects  $n \geq 2$, which involves requires a  significant amount of  higher-order category theory, including weak Kan complexes \cite{weakkan}, quasicategories \citep{quasicats} and $\infty$-categories \citep{kerodon}, which were reviewed above. We pose the problem of causal discovery in simplicial objects in terms of adjoints between categories and simplicial objects. The nerve of a category is a fully faithful right adjoint from a category ${\cal C}$ to an associated simplicial object ${\cal N}_{\cal C}$. Its left adjoint is a destructive functor that only preserves information up to the $2$-simplices. We relate these ideas to existing literature on causal discovery from conditional independence oracles on datasets. 

Many previous studies of causal discovery from interventions, including the {\em conservative} family of intervention targets \citep{DBLP:journals/jmlr/HauserB12}, path queries \citep{DBLP:conf/nips/BelloH18a}, and {\em separating systems} of finite sets or graphs.  \citep{DBLP:conf/uai/Eberhardt08,DBLP:journals/jmlr/HauserB12,DBLP:conf/nips/KocaogluSB17,sepsets,MAOCHENG198415} can all be viewed as imposing a  topology on the sets of variables being intervened.

\subsection{Causal Horns and Conditional Independence} 

We now introduce the concept of causal horns, subsets of simplicial objects, which like the case of PSRs above, represent fragments of the complete simplicial object that is constructed during the process of structure discovery. In the literature on causal discovery, it is common to use conditional independence as a guide to causal discovery.  Conditional independence is a {\em symmetric} property, as we will see below from examining a well-known axiom system for conditional independence in statistics. 

\begin{definition}
 A {\bf Causal Horn} $\Lambda^n_i: \Delta^{op} \rightarrow {\cal C}$ is defined as
 
 \[ (\Lambda^n_i)([m]) = \{ \alpha \in {\bf Hom}_\Delta([m],[n]): [n] \not \subseteq \alpha([m]) \cup \{i \} \} \] 
 \end{definition}
 
\begin{figure}[h] 
 \caption{Causal discovery through the space of all models over $3$ variables shown with their associated imset representations \citep{imset-simplex}. Each candidate DAG defines a {\em causal horn}, a simplicial subobject of the complete simplex on $\Delta^2$, and the process of causal structure discovery can be viewed in terms of the abstract horn filling problem defined above for higher-order categories.  \label{imset-simp-obj}}
\centering
\begin{minipage}{0.7\textwidth}
\includegraphics[scale=0.3]{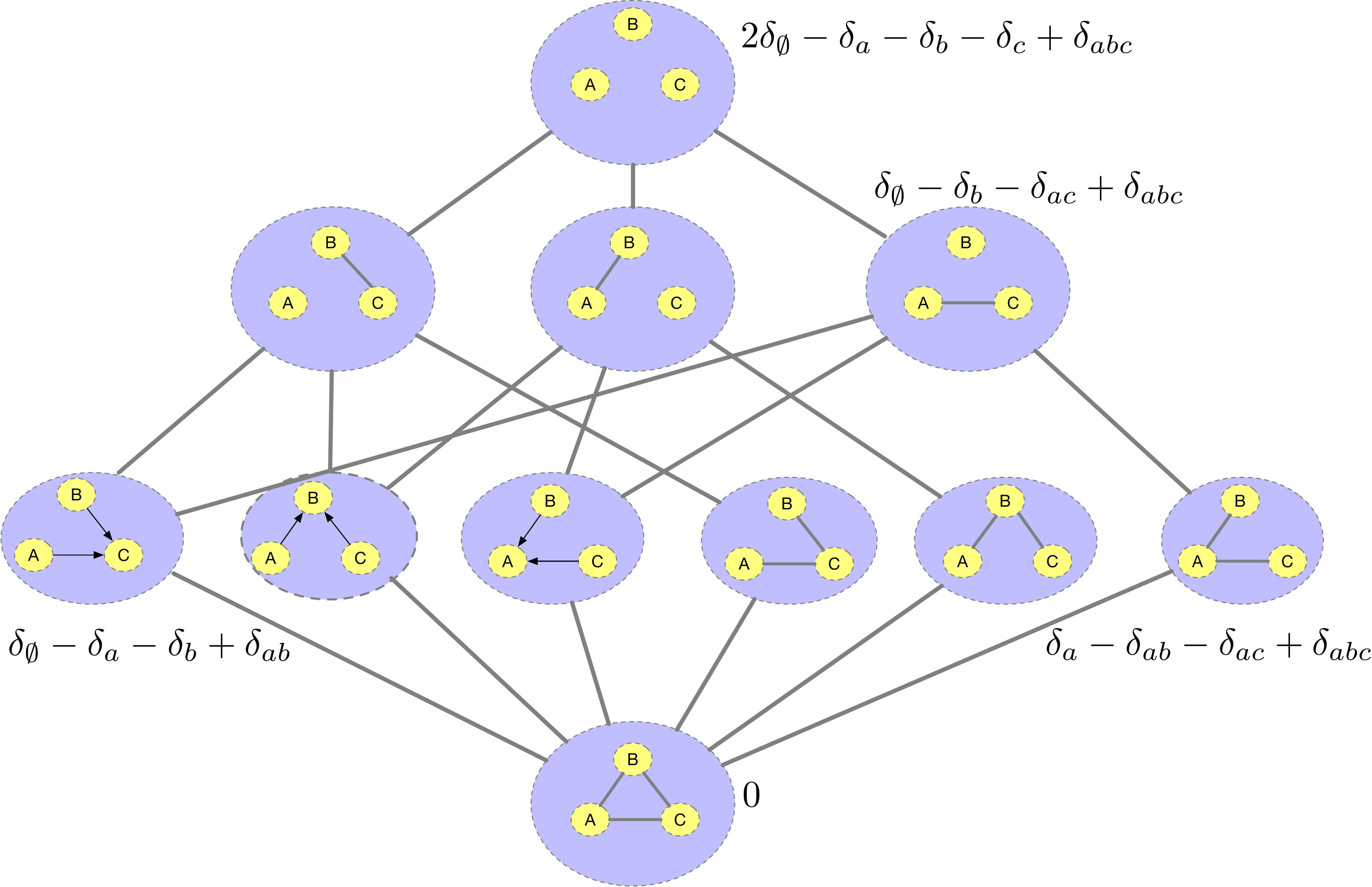}
\end{minipage} 
 \end{figure}

 Intuitively, a causal horn $\Lambda^n_i$ is a fragment of a causal model (see Figure~\ref{imset-simp-obj}) that can be viewed as the simplicial subset that results from removing the interior of the $n$-simplex $\Delta^n$ together with the face opposite its $i$th vertex. Note that ${\cal C}_n$ defines the $n$-simplex associated with the horn of a causal model defined by the category ${\cal C}$  (recall that a simplicial object is a contravariant functor from the category of ordinal numbers $\Delta$ to the category ${\cal C}$ defining causal models.).    

\begin{definition}
\label{separoid}
A {\bf separoid} \citet{DBLP:journals/amai/Dawid01} defines a category over preordered set $({\cal S}, \leq)$, namely $\leq$ is reflexive and transitive, equipped with a {\em ternary} relation $\CI$ on triples $(x,y,z)$, where $x, y, z \in {\cal S}$ satisfy the following properties: 
\begin{itemize}
    \item {\bf S1:} $({\cal S}, \leq)$ is a join semi-lattice. 
    \item {\bf P1:} $x \CI y \ | \ x$
    \item {\bf P2:} $x \CI y \ | \ z \ \ \ \Rightarrow \ \ \ y \CI x \ | z$ 
    \item {\bf P3:} $x \CI  y \ | \ z \ \ \ \mbox{and} \ \ \ w \leq y \ \ \ \Rightarrow \ \ \ x \CI w \ | z$ 
    \item {\bf P4:} $x \CI y \ | \ z \ \ \  \mbox{and} \ \ \ w \leq y \ \ \ \Rightarrow \ \ \ x \CI y \ | \ (z \vee w)$
    \item {\bf P5:} $x \CI y \ | \ z \ \ \ \mbox{and} \ \ \ x \CI w \ | \ (y \vee z) \ \ \ \Rightarrow \ \ \ x \CI (y \vee w) \ | \ z$
\end{itemize}

A {\bf strong separoid} also defines a category. A strong separoid is defined over a lattice ${\cal S}$ has in addition to a join $\vee$, a meet $\wedge$ operation, and satisfies an additional axiom: 

\begin{itemize} 

\item {\bf P6}: If $z \leq y$ and $w \leq y$, then $x \CI y \ | \ z \ \ \ \mbox{and} \ \ \ x \CI y \ | \ w \ \ \ \Rightarrow \ \ \ x \CI y  \ | \ z \ \wedge  \ w$
\end{itemize}

\end{definition} 

It is well known that causal DAG structures can only be recovered up to an equivalence class under observation. For example, the three DAG models $A \rightarrow B \rightarrow C$, $A \leftarrow B \leftarrow C$, and the diverger $A \leftarrow B \rightarrow C$ cannot be discriminated from observations alone, because they define the same conditional independence property $A \CI C | B$. Parameterizing these models  as probability distributions implies that Bayes rule allows inferring $P(B |A)$ from $P(A |B)$, thus these models are equivalent from purely observational data. To discover the exact structure requires making interventions (e.g., intervening on a variable is usually done by clamping its value $X = x$, and thereby deleting any arrows entering the variable from other objects). 
The same problem applies as well to non-graphical representations of conditional independence, such as integer-valued multisets \citep{studeny2010probabilistic}, defined as an integer-valued multiset function $u: \mathbb{Z}^{{\cal P(\mathbb{Z})}} \rightarrow \mathbb{Z}$ from the power set of integers, ${\cal P(\mathbb{Z})}$ to integers $\mathbb{Z}$. An imset is defined over partialy ordered set (poset), defined as a distributive lattice of disjoint (or non-disjoint) subsets of variables. The bottom element is denoted $\emptyset$, and top element represents the complete set of variables $N$. A full discussion of the probabilistic representations induced by imsets is given \citep{studeny2010probabilistic}. We will only focus on the aspects of imsets that relate to its conditional independence structure, and its topological structure as defined by the poset.  A {\em combinatorial} imset is defined as: 

\[ u = \sum_{A \subset N} c_A \delta_A \]

where $c_A$ is an integer, $\delta_A$ is the characteristic function for subset $A$, and $A$ potentially ranges over all subsets of $N$. An {\em elementary} imset is defined over $(a,b \CI A)$, where $a,b$ are singletons, and $A \subset N \setminus \{a, b\}$. A {\em structural} imset is defined as one where the coefficients can be rational numbers. For a general DAG model $G = (V, E)$, an imset in standard form \citep{studeny2010probabilistic} is defined as 

\[ u_G = \delta_V - \delta_\emptyset + \sum_{i \in V} (\delta_{\mbox{{\bf  Pa}}_i} - \delta_{i \cup \mbox{{\bf Pa}}_i}) \] 

\begin{figure}[h] 
 \caption{An illustration of an integer-valued multiset (imset) consisting of a lattice of subsets over three elements for representing conditional independences in DAG models. All three DAG models are represented by the same imset. We can view an imset as a functor from a simplicial object (shown as the dark blue filled triangle) to a left $k$ module. Note that a causal intervention on a DAG  generates a horn of the simplicial object shown in dark blue.  \label{imset}}
\centering
\begin{minipage}{0.5\textwidth}
\includegraphics[scale=0.3]{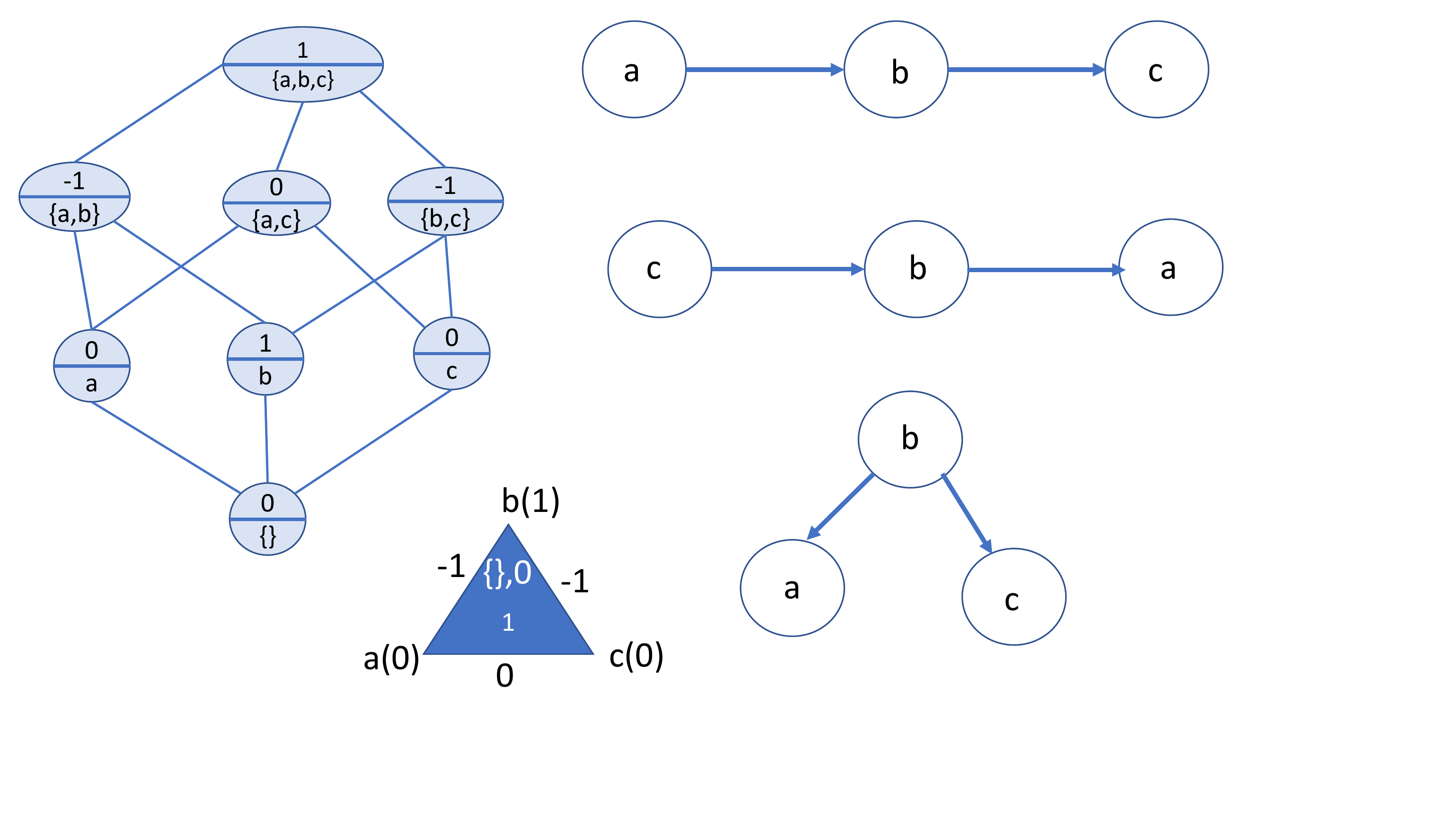}
\end{minipage} 
 \end{figure}

Figure~\ref{imset} shows an example imset for DAG models over three variables, defined by an integer valued function over the lattice of subsets. Each of the three DAG models shown defines exactly the same imset function.  \citet{studeny2010probabilistic} gives a detailed analysis of imsets as a non-graphical representation of conditional independence. 

Figure~\ref{imset-simp-obj} illustrates the space of possible causal DAG models over $3$ variables, with their associated imset representations (see \citep{imset-simplex} for details). Each vertex in this figure shows a potential DAG model. To explain this representation in more detail, the undirected edges represent the essential graph, namely for each edge of the form $a - b$, there are two possible structures that it represents $a \rightarrow b$ and $a \leftarrow b$. These two structures are homotopically equivalent, that is, they define an equivalence class on the space of all models. 

\subsection{Causal Discovery as Adjoint Functors} 

Following the case for PSR structure discovery, we can also formulate  causal discovery in terms of a pair of adjoint functors between a category ${\bf C}$ and a simplicial object $X$. 

\[
        \begin{tikzcd}
            X_n \arrow[r, shift left=1ex, "G"{name=G}] & \C\arrow[l, shift left=.5ex, "F"{name=F}]
            \arrow[phantom, from=F, to=G, , "\scriptscriptstyle\boldsymbol{\top}"].
        \end{tikzcd}
    \]
    
In general, the functor $G$ from a simplicial object $X$ to a category ${\cal C}$ can be lossy. For example, we can define the objects of ${\cal C}$ to be the elements of $X_0$, and the morphisms of ${\cal C}$ as the elements of $f \in X_1$, where $f: a \rightarrow b$, and $d_0 f = a$, and $d_1 f = b$, and $s_0 a, a \in X$ as defining the identity morphisms ${\bf 1}_a$. Composition in this case can be defined as the free algebra defined over elements of $X_1$, subject to the constraints given by elements of $X_2$. For example, if $x \in X_2$, we can impose the requirement that $d_1 x = d_0 x \circ d_2 x$. Such a definition of the left adjoint would be quite lossy because it only preserves the structure of the simplicial object $X$ up to the $2$-simplices. The right adjoint from a category to its associated simplicial object, in contrast, constructs a full and faithful embedding of a category into a simplicial set.  In particular, the  nerve of a category is such a right adjoint. 

Consider the three horns defined below, once again, and let us relate them to the partial DAG structures shown in Figure~\ref{imset-simp-obj}. Notice all three horns are illustrated by candidate DAG structures, for example, the rightmost horn below corresponds to the partial DAG $A \rightarrow C \leftarrow B$, under the mapping $0 \rightarrow A, 1 \rightarrow B, 2 \rightarrow C$. The edge that represents moving from this partial DAG to the complete DAG at the bottom of Figure~\ref{imset-simp-obj} corresponds exactly to the problem of filling the outer horn shown below on the right. 

\begin{center}
 \begin{tikzcd}[column sep=small]
& \{0\}  \arrow[dl] \arrow[dr] & \\
  \{1 \} \arrow[rr, dashed] &                         & \{ 2 \} 
\end{tikzcd} \hskip 0.5 in 
 \begin{tikzcd}[column sep=small]
& \{0\}  \arrow[dl] \arrow[dr, dashed] & \\
  \{1 \} \arrow{rr} &                         & \{ 2 \} 
\end{tikzcd} \hskip 0.5in 
 \begin{tikzcd}[column sep=small]
& \{0\}  \arrow[dl, dashed] \arrow[dr] & \\
  \{1 \} \arrow{rr} &                         & \{ 2 \} 
\end{tikzcd}
\end{center}

Let us consider a more complex case of horn filling over a $3$-simplex $f: \Lambda^3_2 \rightarrow {\cal C}$ for any category ${\cal C}$ that represents a causal model. The ``data" that is used to solve the ``horn filling" problem is as follows: 

\begin{itemize} 

\item There are $4$ objects, represented as $0$-simplices $x_0, x_1, x_2$ and $x_3$. 

\item There are $6$ $1$-simplices, which we can define as potential causal arrows $f_{0,1}, f_{1,2}, f_{2,3}, f_{0,2}, f_{0,3}$ and $f_{1,3}$. Each of these arrows $f_{i,j}$ defines two $0$-faces $x_i$ and $x_j$, the basic units in a causal study. 

\item There are only $3$ $2$-simplices $f_{0,2,3}, f_{0,1,2}$ and $f_{1,2,3}$. As we are dealing with the horn $\Lambda^3_2$, the $2$-simplex $f_{0,1,3}$ is missing. Each of the given $2$-simplices defines a homotopy equivalent composition, for example $f_{0,2,3}: f_{2,3} \circ f_{0,2} \sim f_{0,3}$. 

\end{itemize} 

The problem of horn filling is to deduce the missing $2$-simplex $f_{0,1,3}$ and the missing $3$-simplex $f_{0,1,2,3}$ by defining a homotopy of compositions of homotopies: 

\[ f_{0,1,2,3}: f_{0,2,3} \circ f_{0,1,2} \equiv f_{0,1,3} \circ f_{1,2,3} \]

\begin{definition}
The {\bf nerve} of a causal model ${\cal C}$ is the set of sequences of composable arrows of length $n$, for $n \geq 1$.  Let $N_n({\cal C})$ denote the set of sequences of composable arrows 

\[ \{ C_o \xrightarrow[]{f_1} C_1 \xrightarrow[]{f_2} \ldots \xrightarrow[]{f_n} C_n \ | \ C_i \ \mbox{is an object in} \ {\cal C}, f_i \ \mbox{is a morphism in} \ {\cal C} \} \] 

We can define the causal discovery problem in terms of the right adjoint from the nerve of a causal model to the category. In general, the functor mapping a simplicial object to a category is lossy, because it only preserves properties of the simplicial object up to a certain level. In the case of causal discovery, we can exploit symmetry properties from conditional independence to design this functorial mapping, as was discussed above. 

\[
        \begin{tikzcd}
            {\cal N}_\bullet ({\cal C}) \arrow[r, shift left=1ex, "G"{name=G}] & \C\arrow[l, shift left=.5ex, "F"{name=F}]
            \arrow[phantom, from=F, to=G, , "\scriptscriptstyle\boldsymbol{\top}"].
        \end{tikzcd}
    \]
    
\end{definition}

 The definition of nerve leads to an important topological characterization of any causal category. 

\begin{definition}
The {\bf classifying space} of a causal model ${\cal C}$ is the topological space $|N_\bullet({\cal C})|$. 
\end{definition}

The classifying space of a causal model gives a way to define an algebraic invariant. 

\subsection{The Singular Homology of a Causal Model} 

We will now describe the singular homology of a causal model, using the imset representation as an example shown earlier in Figure~\ref{imset-simp-obj}.   

\begin{definition}
For any topological space $X$ defined by a causal model,  the {\bf singular homology groups} $H_*(X; {\bf Z})$ are defined as the homology groups of a chain complex 

\[ \ldots \xrightarrow[]{\partial} {\bf Z}(\mbox{Sing}_2(X)) \xrightarrow[]{\partial} {\bf Z}(\mbox{Sing}_1(X)) \xrightarrow[]{\partial} {\bf Z}(\mbox{Sing}_0(X)) \] 

where ${\bf Z}(\mbox{Sing}_n(X))$ denotes the free Abelian group generated by the set $\mbox{Sing}_n(X)$ and the differential $\partial$ is defined on the generators by the formula 

\[ \partial (\sigma) = \sum_{i=0}^n (-1)^i d_i \sigma \] 
\end{definition}

Intuitively, a chain complex builds a sequence of vector spaces that can be used to construct an algebraic invariant of a topological space by choosing the left {\bf k} module ${\bf Z}$ to be a vector space. Each differential $\partial$ then becomes a linear transformation whose representation is constructed by modeling its effect on the basis elements in each ${\bf Z}(\mbox{Sing}_n(X))$. 

\begin{example}
Let us illustrate the singular homology groups defined by an integer-valued multiset \cite{studeny2010probabilistic} used to model conditional independence. Imsets over a DAG of three variables $N = \{a, b, c \} $ shown previously as Figure~\ref{imset} can be viewed as a finite discrete topological space. For this topological space $X$, the singular homology groups $H_*(X; {\bf Z})$ are defined as the homology groups of a chain complex 

\[  {\bf Z}(\mbox{Sing}_3(X)) \xrightarrow[]{\partial}  {\bf Z}(\mbox{Sing}_2(X)) \xrightarrow[]{\partial} {\bf Z}(\mbox{Sing}_2(X)) \xrightarrow[]{\partial} {\bf Z}(\mbox{Sing}_1(X)) \xrightarrow[]{\partial} {\bf Z}(\mbox{Sing}_0(X)) \] 

where ${\bf Z}(\mbox{Sing}_i(X))$ denotes the free Abelian group generated by the set $\mbox{Sing}_i(X)$ and the differential $\partial$ is defined on the generators by the formula 

\[ \partial (\sigma) = \sum_{i=0}^4 (-1)^i d_i \sigma \] 

The set $\mbox{Sing}_n(X)$ is the set of all morphisms ${\bf Hom}_{Top}(|\Delta_n|, X)$. For an imset over the three variables $N = \{a, b, c \}$, we can define the singular $n$-simplex $\sigma$ as: 

\[ \sigma: |\Delta^4| \rightarrow X \ \ \mbox{where} \ \ |\Delta^n | = \{t_0, t_1, t_2, t_3 \in [0,1]^4 : t_0 + t_1 + t_2 + t_3 = 1 \} \] 

The $n$-simplex $\sigma$ has a collection of faces denoted as $d_0 \sigma, d_1 \sigma, d_2 \sigma$ and $ d_3 \sigma$.  If we pick the $k$-left module ${\bf Z}$ as the vector space over real numbers $\mathbb{R}$, then the above chain complex represents a sequence of vector spaces that can be used to construct an algebraic invariant of a topological space defined by the integer-valued multiset.  Each differential $\partial$ then becomes a linear transformation whose representation is constructed by modeling its effect on the basis elements in each ${\bf Z}(\mbox{Sing}_n(X))$. An alternate approach to constructing a chain homology for an integer-valued multiset is to use M\"obius inversion to define the chain complex in terms of the nerve of a category (see our recent work on categoroids \citep{categoroids} for details). 
\end{example}

\section{Summary} 

In this paper we presented a unified formalism for  structure discovery in  causal inference and reinforcement learning using the framework of simplicial objects, contravariant functors from the category of ordinal numbers into any category, which form the basis for higher-order category theory. We showed that structure discovery in both causal inference and RL can be defined as filling the horns of simplicial objects, and showed that a number of constructions in higher-order category theory, such as extensions and lifting problems, weak Kan complexes, nerve of a category, and chain complexes could be brought to bear on the design of new algorithms.   Specifically, we show that  fragments of causal models that are equivalent under conditional independence -- defined as causal horns -- as well as fragments of potential tests in a predictive state representation -- defined as predictive horns -- are both special cases of horns of a simplicial object of order $n$,  resulting from the removal of the interior and the face opposite a particular vertex in an $n$-simplex. Causal and predictive-state horns both result from interventions on simplicial objects,  defined as images of a sequence of elementary order-preserving morphisms from the category of ordinal numbers $\Delta$ into an underlying category encoding a universal causal or decision model.  Latent structure discovery in both settings involve the same fundamental mathematical problem of finding extensions of horns of simplicial objects through solving lifting problems in commutative diagrams, and exploiting weak homotopies that define higher-order symmetries. The problem of filling ``inner horns" is solved using quasicategories and weak Kan extensions, whereas filling ``outer" horns requires ideas from $\infty$-category theory. We define the common abstract problem of structure discovery in terms of adjoint functors between a universal causal or decision model category and its simplicial object representation.   In general, the left adjoint functor from a simplicial object $X$ to a category ${\cal C}$ is lossy, preserving only relationships up to a certain order defined by homotopical equivalences. In contrast, the right adjoint defining the nerve of a category constructs a lossless encoding of a category as a simplicial object.


\begin{thebibliography}{47}
\providecommand{\natexlab}[1]{#1}
\providecommand{\url}[1]{\texttt{#1}}
\expandafter\ifx\csname urlstyle\endcsname\relax
  \providecommand{\doi}[1]{doi: #1}\else
  \providecommand{\doi}{doi: \begingroup \urlstyle{rm}\Url}\fi

\bibitem[May(1992)]{may1992simplicial}
J.P. May.
\newblock \emph{Simplicial Objects in Algebraic Topology}.
\newblock Chicago Lectures in Mathematics. University of Chicago Press, 1992.
\newblock ISBN 9780226511818.
\newblock URL \url{https://books.google.com/books?id=QGjwV0gyQnIC}.

\bibitem[Boardman and Vogt(1973)]{weakkan}
M.~Boardman and Rainer Vogt.
\newblock \emph{Homotopy invariant algebraic structures on topological spaces}.
\newblock Springer, Berlin, 1973.

\bibitem[Joyal(2002)]{quasicats}
A.~Joyal.
\newblock Quasi-categories and kan complexes.
\newblock \emph{Journal of Pure and Applied Algebra}, 175\penalty0
  (1):\penalty0 207--222, 2002.
\newblock ISSN 0022-4049.
\newblock \doi{https://doi.org/10.1016/S0022-4049(02)00135-4}.
\newblock URL
  \url{https://www.sciencedirect.com/science/article/pii/S0022404902001354}.
\newblock Special Volume celebrating the 70th birthday of Professor Max Kelly.

\bibitem[Lurie(2022)]{kerodon}
Jacob Lurie.
\newblock Kerodon.
\newblock \url{https://kerodon.net}, 2022.

\bibitem[Pearl(2009{\natexlab{a}})]{pearl:causalitybook}
Judea Pearl.
\newblock \emph{Causality: Models, Reasoning and Inference}.
\newblock Cambridge University Press, USA, 2nd edition, 2009{\natexlab{a}}.
\newblock ISBN 052189560X.

\bibitem[Imbens and Rubin(2015)]{rubin-book}
Guido~W. Imbens and Donald~B. Rubin.
\newblock \emph{Causal Inference for Statistics, Social, and Biomedical
  Sciences: An Introduction}.
\newblock Cambridge University Press, USA, 2015.
\newblock ISBN 0521885884.

\bibitem[Spirtes et~al.(2000)Spirtes, Glymour, and Scheines]{spirtes:book}
Peter Spirtes, Clark Glymour, and Richard Scheines.
\newblock \emph{Causation, Prediction, and Search, Second Edition}.
\newblock Adaptive computation and machine learning. {MIT} Press, 2000.
\newblock ISBN 978-0-262-19440-2.

\bibitem[Singh et~al.(2004)Singh, James, and Rudary]{singh-uai04}
Satinder~P. Singh, Michael~R. James, and Matthew~R. Rudary.
\newblock Predictive state representations: {A} new theory for modeling
  dynamical systems.
\newblock In David~Maxwell Chickering and Joseph~Y. Halpern, editors,
  \emph{{UAI} '04, Proceedings of the 20th Conference in Uncertainty in
  Artificial Intelligence, Banff, Canada, July 7-11, 2004}, pages 512--518.
  {AUAI} Press, 2004.

\bibitem[Sutton and Barto(1998)]{DBLP:books/lib/SuttonB98}
Richard~S. Sutton and Andrew~G. Barto.
\newblock \emph{Reinforcement learning - an introduction}.
\newblock Adaptive computation and machine learning. {MIT} Press, 1998.
\newblock ISBN 978-0-262-19398-6.
\newblock URL \url{https://www.worldcat.org/oclc/37293240}.

\bibitem[Van~Overschee and De~Moor(1996)]{overschee}
Peter Van~Overschee and Bart De~Moor.
\newblock \emph{Subspace identification for linear systems. Theory,
  implementation, applications. Incl. 1 disk}, volume xiv, pages xiv + 254.
\newblock Springer, 01 1996.
\newblock ISBN 0-7923-9717-7.
\newblock \doi{10.1007/978-1-4613-0465-4}.

\bibitem[Chomsky and Sch\"utzenberger(1963)]{CHOMSKY1963118}
N.~Chomsky and M.P. Sch\"utzenberger.
\newblock The algebraic theory of context-free languages*.
\newblock In P.~Braffort and D.~Hirschberg, editors, \emph{Computer Programming
  and Formal Systems}, volume~35 of \emph{Studies in Logic and the Foundations
  of Mathematics}, pages 118--161. Elsevier, 1963.
\newblock \doi{https://doi.org/10.1016/S0049-237X(08)72023-8}.
\newblock URL
  \url{https://www.sciencedirect.com/science/article/pii/S0049237X08720238}.

\bibitem[Give'on and Arbib(1968)]{GIVEON1968346}
Y.~Give'on and M.A. Arbib.
\newblock Algebra automata ii: The categorical framework for dynamic analysis.
\newblock \emph{Information and Control}, 12\penalty0 (4):\penalty0 346--370,
  1968.
\newblock ISSN 0019-9958.
\newblock \doi{https://doi.org/10.1016/S0019-9958(68)90381-1}.
\newblock URL
  \url{https://www.sciencedirect.com/science/article/pii/S0019995868903811}.

\bibitem[Pearl(1989)]{pearl:bnets-book}
Judea Pearl.
\newblock \emph{Probabilistic reasoning in intelligent systems - networks of
  plausible inference}.
\newblock Morgan Kaufmann series in representation and reasoning. Morgan
  Kaufmann, 1989.

\bibitem[Lauritzen and Richardson(2002)]{lauritzen:chain}
Steffen~L. Lauritzen and Thomas~S. Richardson.
\newblock Chain graph models and their causal interpretations.
\newblock \emph{Journal of the Royal Statistical Society: Series B (Statistical
  Methodology)}, 64\penalty0 (3):\penalty0 321--348, 2002.
\newblock \doi{https://doi.org/10.1111/1467-9868.00340}.
\newblock URL
  \url{https://rss.onlinelibrary.wiley.com/doi/abs/10.1111/1467-9868.00340}.

\bibitem[Forré and Mooij(2017)]{hedge}
Patrick Forré and Joris~M. Mooij.
\newblock Markov properties for graphical models with cycles and latent
  variables, 2017.

\bibitem[Evans(2018)]{mdag}
Robin~J. Evans.
\newblock {Margins of discrete Bayesian networks}.
\newblock \emph{The Annals of Statistics}, 46\penalty0 (6A):\penalty0 2623 --
  2656, 2018.
\newblock \doi{10.1214/17-AOS1631}.
\newblock URL \url{https://doi.org/10.1214/17-AOS1631}.

\bibitem[Dawid(2001)]{DBLP:journals/amai/Dawid01}
A.~Philip Dawid.
\newblock Separoids: {A} mathematical framework for conditional independence
  and irrelevance.
\newblock \emph{Ann. Math. Artif. Intell.}, 32\penalty0 (1-4):\penalty0
  335--372, 2001.
\newblock \doi{10.1023/A:1016734104787}.
\newblock URL \url{https://doi.org/10.1023/A:1016734104787}.

\bibitem[Studený et~al.(2010{\natexlab{a}})Studený, Vomlel, and
  Hemmecke]{STUDENY2010573}
Milan Studený, Jiří Vomlel, and Raymond Hemmecke.
\newblock A geometric view on learning bayesian network structures.
\newblock \emph{International Journal of Approximate Reasoning}, 51\penalty0
  (5):\penalty0 573--586, 2010{\natexlab{a}}.
\newblock ISSN 0888-613X.
\newblock \doi{https://doi.org/10.1016/j.ijar.2010.01.014}.
\newblock URL
  \url{https://www.sciencedirect.com/science/article/pii/S0888613X10000216}.
\newblock PGM-2008.

\bibitem[Mahadevan(2022{\natexlab{a}})]{sm:uc}
Sridhar Mahadevan.
\newblock On the universality of diagrams for causal inference and the causal
  reproducing property, 2022{\natexlab{a}}.
\newblock URL \url{https://arxiv.org/abs/2207.02917}.

\bibitem[Mahadevan(2021{\natexlab{a}})]{sm:udm}
Sridhar Mahadevan.
\newblock Universal decision models.
\newblock \emph{CoRR}, abs/2110.15431, 2021{\natexlab{a}}.
\newblock URL \url{https://arxiv.org/abs/2110.15431}.

\bibitem[May(1999)]{may1999concise}
J.P. May.
\newblock \emph{A Concise Course in Algebraic Topology}.
\newblock Chicago Lectures in Mathematics. University of Chicago Press, 1999.
\newblock ISBN 9780226511832.
\newblock URL \url{https://books.google.com/books?id=g8SG03R1bpgC}.

\bibitem[MacLane(1971)]{maclane:71}
Saunders MacLane.
\newblock \emph{Categories for the Working Mathematician}.
\newblock Springer-Verlag, New York, 1971.
\newblock Graduate Texts in Mathematics, Vol. 5.

\bibitem[Richter(2020)]{richter2020categories}
B.~Richter.
\newblock \emph{From Categories to Homotopy Theory}.
\newblock Cambridge Studies in Advanced Mathematics. Cambridge University
  Press, 2020.
\newblock ISBN 9781108479622.
\newblock URL \url{https://books.google.com/books?id=pnzUDwAAQBAJ}.

\bibitem[Puterman(1994)]{DBLP:books/wi/Puterman94}
Martin~L. Puterman.
\newblock \emph{Markov Decision Processes: Discrete Stochastic Dynamic
  Programming}.
\newblock Wiley Series in Probability and Statistics. Wiley, 1994.
\newblock ISBN 978-0-47161977-2.
\newblock \doi{10.1002/9780470316887}.
\newblock URL \url{https://doi.org/10.1002/9780470316887}.

\bibitem[Ravindran and Barto(2003)]{DBLP:conf/ijcai/RavindranB03}
Balaraman Ravindran and Andrew~G. Barto.
\newblock {SMDP} homomorphisms: An algebraic approach to abstraction in
  semi-markov decision processes.
\newblock In Georg Gottlob and Toby Walsh, editors, \emph{IJCAI-03, Proceedings
  of the Eighteenth International Joint Conference on Artificial Intelligence,
  Acapulco, Mexico, August 9-15, 2003}, pages 1011--1018. Morgan Kaufmann,
  2003.
\newblock URL \url{http://ijcai.org/Proceedings/03/Papers/145.pdf}.

\bibitem[Fong and Spivak(2018)]{fong2018seven}
Brendan Fong and David~I Spivak.
\newblock Seven sketches in compositionality: An invitation to applied category
  theory, 2018.
\newblock URL \url{http://arxiv.org/abs/1803.05316}.
\newblock cite arxiv:1803.05316Comment: 341+xii pages.

\bibitem[Bradley et~al.(2022)Bradley, Terilla, and Vlassopoulos]{Bradley_2022}
Tai-Danae Bradley, John Terilla, and Yiannis Vlassopoulos.
\newblock An enriched category theory of language: From syntax to semantics.
\newblock \emph{La Matematica}, mar 2022.
\newblock \doi{10.1007/s44007-022-00021-2}.
\newblock URL \url{https://doi.org/10.1007\%2Fs44007-022-00021-2}.

\bibitem[Thon and Jaeger(2015)]{DBLP:journals/jmlr/ThonJ15}
Michael~R. Thon and Herbert Jaeger.
\newblock Links between multiplicity automata, observable operator models and
  predictive state representations: a unified learning framework.
\newblock \emph{J. Mach. Learn. Res.}, 16:\penalty0 103--147, 2015.
\newblock URL \url{http://dl.acm.org/citation.cfm?id=2789276}.

\bibitem[Soni and Singh(2007)]{DBLP:conf/aaai/SoniS07}
Vishal Soni and Satinder~P. Singh.
\newblock Abstraction in predictive state representations.
\newblock In \emph{Proceedings of the Twenty-Second {AAAI} Conference on
  Artificial Intelligence, July 22-26, 2007, Vancouver, British Columbia,
  Canada}, pages 639--644. {AAAI} Press, 2007.
\newblock URL \url{http://www.aaai.org/Library/AAAI/2007/aaai07-101.php}.

\bibitem[Janzing et~al.(2013)Janzing, Balduzzi, Grosse-Wentrup, and
  Schölkopf]{janzing}
Dominik Janzing, David Balduzzi, Moritz Grosse-Wentrup, and Bernhard
  Schölkopf.
\newblock {Quantifying causal influences}.
\newblock \emph{The Annals of Statistics}, 41\penalty0 (5):\penalty0 2324 --
  2358, 2013.
\newblock \doi{10.1214/13-AOS1145}.
\newblock URL \url{https://doi.org/10.1214/13-AOS1145}.

\bibitem[Zigler and Papadogeorgou(2018)]{zigler2018bipartite}
Corwin~M. Zigler and Georgia Papadogeorgou.
\newblock Bipartite causal inference with interference, 2018.

\bibitem[Adam and Dahleh(2019)]{adams}
Elie~M. Adam and Munther~A. Dahleh.
\newblock Generativity and interactional effects: an overview, 2019.
\newblock URL \url{https://arxiv.org/abs/1911.10406}.

\bibitem[Nagurney(1999)]{nagurney:vibook}
A.~Nagurney.
\newblock \emph{Network Economics: A Variational Inequality Approach}.
\newblock Kluwer Academic Press, 1999.

\bibitem[Mahadevan(2021{\natexlab{b}})]{sm:causal-network-econ}
Sridhar Mahadevan.
\newblock Causal inference in network economics.
\newblock \emph{CoRR}, abs/2109.11344, 2021{\natexlab{b}}.
\newblock URL \url{https://arxiv.org/abs/2109.11344}.

\bibitem[Pearl(2009{\natexlab{b}})]{pearl-book}
Judea Pearl.
\newblock \emph{Causality: Models, Reasoning and Inference}.
\newblock Cambridge University Press, USA, 2nd edition, 2009{\natexlab{b}}.
\newblock ISBN 052189560X.

\bibitem[Witsenhausen(1975)]{witsenhausen:1975}
H.~S. Witsenhausen.
\newblock The intrinsic model for discrete stochastic control: Some open
  problems.
\newblock In A.~Bensoussan and J.~L. Lions, editors, \emph{Control Theory,
  Numerical Methods and Computer Systems Modelling}, pages 322--335, Berlin,
  Heidelberg, 1975. Springer Berlin Heidelberg.
\newblock ISBN 978-3-642-46317-4.

\bibitem[Heymann et~al.(2021)Heymann, de~Lara, and Chancelier]{heymann:if}
Benjamin Heymann, Michel de~Lara, and Jean-Philippe Chancelier.
\newblock Causal inference theory with information dependency models, 2021.
\newblock URL \url{https://arxiv.org/abs/2108.03099}.

\bibitem[Mahadevan(2021{\natexlab{c}})]{sm:homotopy}
Sridhar Mahadevan.
\newblock Causal homotopy, 2021{\natexlab{c}}.
\newblock URL \url{https://arxiv.org/abs/2112.01847}.

\bibitem[Hauser and B{\"{u}}hlmann(2012)]{DBLP:journals/jmlr/HauserB12}
Alain Hauser and Peter B{\"{u}}hlmann.
\newblock Characterization and greedy learning of interventional markov
  equivalence classes of directed acyclic graphs.
\newblock \emph{J. Mach. Learn. Res.}, 13:\penalty0 2409--2464, 2012.
\newblock URL \url{http://dl.acm.org/citation.cfm?id=2503320}.

\bibitem[Bello and Honorio(2018)]{DBLP:conf/nips/BelloH18a}
Kevin Bello and Jean Honorio.
\newblock Computationally and statistically efficient learning of causal bayes
  nets using path queries.
\newblock In Samy Bengio, Hanna~M. Wallach, Hugo Larochelle, Kristen Grauman,
  Nicol{\`{o}} Cesa{-}Bianchi, and Roman Garnett, editors, \emph{Advances in
  Neural Information Processing Systems 31: Annual Conference on Neural
  Information Processing Systems 2018, NeurIPS 2018, December 3-8, 2018,
  Montr{\'{e}}al, Canada}, pages 10954--10964, 2018.
\newblock URL
  \url{https://proceedings.neurips.cc/paper/2018/hash/a0b45d1bb84fe1bedbb8449764c4d5d5-Abstract.html}.

\bibitem[Eberhardt(2008)]{DBLP:conf/uai/Eberhardt08}
Frederick Eberhardt.
\newblock Almost optimal intervention sets for causal discovery.
\newblock In David~A. McAllester and Petri Myllym{\"{a}}ki, editors,
  \emph{{UAI} 2008, Proceedings of the 24th Conference in Uncertainty in
  Artificial Intelligence, Helsinki, Finland, July 9-12, 2008}, pages 161--168.
  {AUAI} Press, 2008.
\newblock URL
  \url{https://dslpitt.org/uai/displayArticleDetails.jsp?mmnu=1\&smnu=2\&article\_id=1948\&proceeding\_id=24}.

\bibitem[Kocaoglu et~al.(2017)Kocaoglu, Shanmugam, and
  Bareinboim]{DBLP:conf/nips/KocaogluSB17}
Murat Kocaoglu, Karthikeyan Shanmugam, and Elias Bareinboim.
\newblock Experimental design for learning causal graphs with latent variables.
\newblock In Isabelle Guyon, Ulrike von Luxburg, Samy Bengio, Hanna~M. Wallach,
  Rob Fergus, S.~V.~N. Vishwanathan, and Roman Garnett, editors, \emph{Advances
  in Neural Information Processing Systems 30: Annual Conference on Neural
  Information Processing Systems 2017, December 4-9, 2017, Long Beach, CA,
  {USA}}, pages 7018--7028, 2017.
\newblock URL
  \url{https://proceedings.neurips.cc/paper/2017/hash/291d43c696d8c3704cdbe0a72ade5f6c-Abstract.html}.

\bibitem[Katona(1966)]{sepsets}
Gyula Katona.
\newblock {On separating systems of a finite set}.
\newblock \emph{Journal of Combinatorial Theory}, 2\penalty0 (1):\penalty0
  174--194, 1966.

\bibitem[Mao-cheng(1984)]{MAOCHENG198415}
CAI Mao-cheng.
\newblock On separating systems of graphs.
\newblock \emph{Discrete Mathematics}, 49\penalty0 (1):\penalty0 15--20, 1984.
\newblock ISSN 0012-365X.
\newblock \doi{https://doi.org/10.1016/0012-365X(84)90146-8}.
\newblock URL
  \url{https://www.sciencedirect.com/science/article/pii/0012365X84901468}.

\bibitem[Studeny(2010)]{studeny2010probabilistic}
M.~Studeny.
\newblock \emph{Probabilistic Conditional Independence Structures}.
\newblock Information Science and Statistics. Springer London, 2010.
\newblock ISBN 9781849969482.
\newblock URL \url{https://books.google.com.gi/books?id=bGFRcgAACAAJ}.

\bibitem[Studený et~al.(2010{\natexlab{b}})Studený, Vomlel, and
  Hemmecke]{imset-simplex}
Milan Studený, Jiří Vomlel, and Raymond Hemmecke.
\newblock A geometric view on learning bayesian network structures.
\newblock \emph{International Journal of Approximate Reasoning}, 51:\penalty0
  573--586, 06 2010{\natexlab{b}}.
\newblock \doi{10.1016/j.ijar.2010.01.014}.

\bibitem[Mahadevan(2022{\natexlab{b}})]{categoroids}
Sridhar Mahadevan.
\newblock Categoroids: Universal conditional independence, 2022{\natexlab{b}}.
\newblock URL \url{https://arxiv.org/abs/2208.11077}.

\end{thebibliography}

\end{document}